\documentclass[12pt]{article}
\usepackage{graphicx} 
\usepackage{amsfonts}
\usepackage{amsmath}
\usepackage{subcaption}
\usepackage{booktabs}
\usepackage{booktabs}
\usepackage{adjustbox}
\usepackage{hyperref}
\usepackage{multirow}

\usepackage{comment}
\usepackage[symbol]{footmisc}
\providecommand{\keywords}[1]{\textbf{\textit{Index terms---}} #1}
\hypersetup{pdfborder={0 0 0}}

\usepackage{xcolor}

\usepackage{hyphenat}
\title{AI-driven Generation of MALDI-TOF MS for Microbial Characterization}

\author{
  Lucía Schmidt-Santiago$^{1,}$\thanks{Corresponding author (e-mail: lschmidt@pa.uc3m.es)} \and
  David Rodríguez-Temporal$^{2}$ \and
  Carlos Sevilla-Salcedo$^{1}$ \and
  Vanessa G\'{o}mez-Verdejo$^{1,2}$
}

\date{
\footnotesize
  $^{1}$ Department of Signal Theory and Communications, Universidad Carlos III de Madrid, Leganés, 28911, Spain \\
  $^{2}$ Instituto de Investigación Sanitaria Gregorio Marañón (IiSGM), Madrid, 28009 Spain
}

\begin{document}

\maketitle



\begin{abstract}
\noindent
Matrix-Assisted Laser Desorption/Ionization Time-of-Flight Mass Spectrometry (MALDI-TOF MS) has become a cornerstone technology in clinical microbiology, enabling rapid and accurate microbial identification.  
However, the development of data-driven diagnostic models remains limited by the lack of sufficiently large, balanced, and standardized spectral datasets.  
This study investigates the use of deep generative models to synthesize realistic MALDI-TOF MS spectra, aiming to overcome data scarcity and support the development of robust machine learning tools in microbiology.  

We adapt and evaluate three generative models, Variational Autoencoders (MALDIVAEs), Generative Adversarial Networks (MALDIGANs), and Denoising Diffusion Probabilistic Model (MALDIffusion), for the conditional generation of microbial spectra guided by species labels. Generation is conditioned on species labels, and spectral fidelity and diversity are assessed using diverse metrics.


Our experiments show that synthetic data generated by MALDIVAE, MALDIGAN, and MALDIffusion are statistically and diagnostically comparable to real measurements, enabling classifiers trained exclusively on synthetic samples to reach performance levels similar to those trained on real data. While all models faithfully reproduce the peak structure and variability of MALDI-TOF spectra, MALDIffusion obtains this fidelity at a substantially higher computational cost, and MALDIGAN shows competitive but slightly less stable behaviour. In contrast, MALDIVAE offers the most favorable balance between realism, stability, and efficiency. Furthermore, augmenting minority species with synthetic spectra markedly improves classification accuracy, effectively mitigating class imbalance and domain mismatch without compromising the authenticity of the generated data.
\end{abstract}



\section{Introduction}

Few technologies have transformed clinical microbiology as profoundly as Matrix-Assisted Laser Desorption/Ionization Time-of-Flight Mass Spectrometry (MALDI-TOF MS), which enables species-level identification of microorganisms within minutes \cite{dhiman2011performance, patel2013cmr,angeletti2017review}. Since its introduction into clinical diagnostics, MALDI-TOF MS has largely replaced phenotypic and biochemical tests, providing fast and reproducible spectral profiles that support routine hospital workflows and facilitate large-scale surveillance of infectious diseases. However, despite its robustness, MALDI-TOF MS spectra remain sensitive to multiple sources of variability, including differences in culture conditions, extraction protocols, and instrument calibration, that can affect reproducibility and downstream computational analyses \cite{popovic2021msr, hrabak2013resistance}. Moreover, each spectrum is a high-dimensional signal composed of thousands of ion-intensity measurements across the m/z range, encoding rich proteomic information but also posing significant challenges for computational modeling due to its intrinsic dimensionality and noise structure.


In recent years, the combination of MALDI-TOF MS with Machine Learning (ML) techniques has emerged as a promising strategy to enhance microbial identification, detect antimicrobial resistance, and infer phenotypic traits directly from spectral data \cite{weis2020machine,schmidt-santiago2025ml}. Most existing approaches, however, rely on traditional ML algorithms such as Random Forests (RF), Support Vector Machines (SVMs), or Gradient Boosted Trees (e.g., XGBoost) \cite{schmidt-santiago2025ml}. Deep Learning (DL) methods, on the other hand, offer the potential to learn generic and transferable representations of spectra, capturing intrinsic spectral characteristics that are robust to acquisition variability and experimental noise \cite{Nguyen2024, deWaele2025pre}. Yet, the development of such models has been limited by the scarcity of large, publicly available MALDI-TOF MS datasets suitable for pretraining.

Recent initiatives such as DRIAMS \cite{weigand2020driams} and MARISMa \cite{marisma2025dataset} have achieved significant advancements in addressing this challenge. DRIAMS aggregates more than 130,000 unique spectra across multiple hospitals and instruments, while MARISMa provides over 200,000 unique spectra encompassing diverse microbial species and resistance profiles. Despite their size, both datasets remain highly imbalanced; common pathogens such as \textit{Escherichia coli} dominate, while rare or emerging species are underrepresented. This imbalance poses a significant challenge for training advanced DL models and motivates the exploration of methods capable of generating realistic synthetic spectra to improve poorly represented classes.

In parallel, deep generative modeling has achieved remarkable success in other domains of Artificial Intelligence (AI), particularly in text \cite{brown2020language} and computer vision \cite{karras2019style}. 
Models such as Variational Autoencoders (VAEs) \cite{kingma2013auto}, Generative Adversarial Networks (GANs) \cite{goodfellow2014gan,mirza2014cgan, odena2017conditional}, and Denoising Diffusion Probabilistic Models (DDPMs) \cite{ho2020ddpm,song2020sde} have enabled high-fidelity data synthesis and realistic augmentation of complex datasets. These approaches have proven highly effective for data augmentation, style transfer, and representation learning, offering a way to simulate realistic variability and mitigate data scarcity~\cite{shorten2019survey,mittal2023diffusion}.

In this work, we investigate the use of deep generative models to synthesize MALDI-TOF MS spectra for microbiological applications. We adapt and evaluate three representative generative models: VAEs, GANs, and DDPMs, focusing on their conditional formulations to enable species-guided spectral generation. We analyze their capacity to produce realistic and diverse synthetic spectra and discuss their potential for addressing key challenges in clinical microbiology, such as data scarcity and class imbalance.

Specifically, the main contributions of this study are as follows:
\begin{itemize}
    \item We describe how to adapt deep generative models, VAEs, GANs, and DDPMs, to the specific characteristics of MALDI-TOF mass spectra. 
    \item We propose an evaluation methodology that combines distributional and structural similarity metrics to jointly assess the fidelity, diversity, and novelty of generated spectra.
    \item We analyze synthetic data generation across models, assessing consistency and variability metrics to determine which approach most effectively reproduces the distribution of real spectra while maintaining realistic diversity.
    \item We demonstrate that synthetic spectra generated by deep generative models can be used interchangeably with real data for classification tasks, achieving comparable performance and confirming their validity as realistic surrogates.
    \item We evaluate the practical potential of synthetic spectra for improving classification performance in underrepresented microbial species, showing that generative augmentation enhances recognition accuracy in data-limited scenarios.
\end{itemize}

To the best of our knowledge, this work constitutes the first demonstration of deep generative AI applied to the synthesis of MALDI-TOF MS, providing a foundation for future research on synthetic data generation and model robustness in clinical microbiology.

The remainder of this paper is organized as follows: Section~\ref{materials_methods} describes the datasets, preprocessing steps, and the generative modeling approaches considered. Section~\ref{results} presents the experimental setup, the evaluation metrics, and the main findings, including a detailed analysis of the advantages, limitations, and practical implications of each generative approach, as well as its potential in a real-world scenario through the classification of underrepresented microbial species using synthetically augmented data. Finally, Section~\ref{sec:conclusions} summarizes the main conclusions and outlines promising directions for future work.

\section{Materials and Methods}
This section introduces the two datasets considered in this study, together with the experimental partitions defined to evaluate our models under different conditions. We further describe the preprocessing pipeline developed for the MALDI-TOF MS spectra, ensuring consistent and comparable inputs across experiments. Finally, we present the proposed generative approaches and discuss the specific adaptations implemented to tailor each model to the characteristics of the data.

\label{materials_methods}
\subsection{Databases description}


MALDI-TOF MS data consist of one-dimensional spectra representing ion intensities as a function of the mass-to-charge ratio (m/z). Each spectrum captures the distribution of abundant protein peaks characteristic of a microbial isolate, typically spanning the range 2,000–20,000 Da and discretized into a vector of fixed length after preprocessing.

To compile a representative collection of clinically relevant spectra, data were drawn from two complementary sources:

\begin{itemize}
\item The \textbf{Database of Resistance Information on Antimicrobials and MALDI-TOF Mass Spectra (DRIAMS)}\cite{weis2021driams}, a large multi-center dataset comprising over 130,000 unique MALDI-TOF MS, including over 800 distinct microbial species. The dataset includes spectra acquired using the Microflex Biotyper System (Bruker Daltonics) across four hospitals in Switzerland (DRIAMS-A to D) between 2015 and 2018.
\item The \textbf{MALDI-TOF Acquisition Repository of Infectious Species at Marañón (MARISMa)}\cite{schmidt2025marisma}, a single-center dataset containing over 200,000 unique spectra collected between 2018 and 2024 at the Hospital General Universitario Gregorio Marañón (Madrid, Spain). Spectra were acquired using the Microflex Biotyper System (Bruker Daltonics) and cover more than 1,000 different microbial species. 
\end{itemize}

From the wide range of microbial species available in the combined DRIAMS and MARISMa datasets, we selected ten representative species for this study. Six of them were chosen based on their high number of available spectra and their major clinical relevance, since these species represent some of the most prevalent and clinically significant pathogens encountered in hospital microbiology, encompassing agents of both community- and hospital-acquired infections. These include both Gram-positive, \textit{Staphylococcus aureus} (\textit{S.aureus}) and \textit{Enterococcus faecium} (\textit{E. faecium}), and Gram-negative bacteria, \textit{Escherichia coli} (\textit{E. coli}), \textit{Klebsiella pneumoniae} (\textit{K. pneumoniae}), \textit{Pseudomonas aeruginosa} (\textit{P. aeruginosa}), and the \textit{Enterobacter cloacae} (\textit{E. cloacae}) complex. 
The \textit{E. cloacae} complex—a group of closely related species including \textit{E. cloacae}, \textit{E. hormaechei}, \textit{E. asburiae}, \textit{E. kobei}, and \textit{E. ludwigii}—is particularly relevant due to its frequent association with nosocomial infections such as bloodstream and respiratory tract infections. 


In addition to the main species included in the study, we incorporated \textit{Staphylococcus saprophyticus} (\textit{S. saprophyticus}) to evaluate the ability of the generative models to synthesize realistic spectra for minority classes and to assess their utility in improving classifier performance under data-limited conditions. \textit{S. saprophyticus} is a Gram-positive uropathogen and a common cause of uncomplicated urinary tract infections, particularly in young women. While generally responsive to initial antibiotics, occasional resistance has been observed, highlighting the need of accurate identification. In our dataset, \textit{S. saprophyticus} is substantially underrepresented (its number of spectra is less than one third of the samples of the least abundant main species) yet it retains enough observations to support meaningful training and evaluation. This makes it a suitable test case for assessing the benefits of generative augmentation under realistic minority-class constraints.

To emulate a realistic, time-aware scenario while capturing variability across institutions, the dataset was organized into four distinct partitions, ensuring independent train, validation, test and out-of-distribution (OOD) subsets. Specifically, the train partition includes spectra from the first four years of MARISMa (2018--2022) together with DRIAMS hospitals A and B, representing the largest and most diverse portion of historically available data used to train the generative models. The validation set consists of MARISMa spectra acquired in 2023, providing a temporally adjacent but non-overlapping subset used for model tuning. The test partition contains MARISMa data from 2024, forming the most recent temporal block, while the OOD (OOD) partition comprises the DRIAMS cohort C (OOD-C) and cohort D (OOD-D), which originate from different hospitals and acquisition periods. Together, these chronological and institution-based splits enable multiple experimental configurations and reflect realistic scenarios in which model generalization capability will be evaluated across time and acquisition centers.
A detailed breakdown of samples per species in each subset is provided in Table~\ref{tab:selected_species}.

\begin{table}[th]
\centering
\caption{Selected bacterial species and their prevalence in the data splits used.}
\begin{adjustbox}{max width=\textwidth}
    \begin{tabular}{lccccc}
        \toprule
        \textbf{Species} & \textbf{Train} & \textbf{Val} & \textbf{Test} & \textbf{OOD-C/D} & \textbf{Total} \\
        \midrule
        \textit{Escherichia coli}              & 24,224 & 3,133 & 2,074 & \phantom{0}927/2,013 & 32,371 \\
        \textit{Staphylococcus aureus}         & 20,197 & 2,931 & 2,538 & \phantom{0}738/2,174 & 28,578 \\
        \textit{Klebsiella pneumoniae}         & 16,257 & 3,447 & 3,161 & \phantom{0}366/2,151 & 25,382 \\
        \textit{Pseudomonas aeruginosa}        & 14,647 & 2,357 & 2,165 & \phantom{0}357/362\phantom{0}   & 19,888 \\
        \textit{Enterobacter cloacae complex}  & 3,863  & 229   & 160   & \phantom{0}196/437\phantom{0}  & 4,885 \\
        \textit{Enterococcus faecium}          & 2,871  & 404   & 1,313 & \phantom{00}93/171\phantom{0}  & 4,852 \\
        \midrule
        \textit{Staphylococcus saprophyticus}  & 716    & 172   & 168   & \phantom{000}7/200\phantom{0}    & 1,263 \\
        \midrule
        \textbf{Total}                         & 82,775 & 12,673 & 11,579 & 2,684/7,508 & 117,219 \\
        \bottomrule    
    \end{tabular}
\end{adjustbox}
\label{tab:selected_species}
\end{table}


To standardize spectra across datasets, all spectra were processed using a unified preprocessing pipeline \cite{schmidt-santiago2025ml}. The procedure consisted of the following steps.

\begin{enumerate}
    \item Variance stabilization using square root transformation.
    \item Smoothing using the Savitzky-Golay filter with a half-window of 10 to suppress high-frequency noise while preserving peak shapes.
    \item Baseline correction was performed using the Statistics-sensitive Non-linear Iterative Peak-clipping (SNIP) algorithm with 20 iterations for removing low-frequency background signals.
    \item Standard-deviation thresholding by removing residual background noise setting all intensity values below a data-driven threshold to zero.
    \item Trimming spectra to the informative mass range between 2,000 and 20,000Da.
    \item Binning at 3Da intervals, producing uniform-length feature vectors across samples.
    \item Normalization of intensity values to the range [0, 1] for ensuring scale consistency.
\end{enumerate}


After preprocessing, each spectrum was represented as a vector of approximately 6,000 features, each corresponding to a discrete m/z bin. The relative intensities and spatial distribution of peaks reflect the abundance of specific ribosomal and housekeeping proteins, which serve as molecular identifications of each microbial species. The resulting feature space is therefore high-dimensional and sparse, with informative signals localized around a limited number of discriminative peaks contained in a background of low-intensity noise.
\subsection{Generative Modeling of MALDI-TOF MS}

Generative models are a class of ML methods designed to learn the underlying probability distribution of complex data and to generate new synthetic samples that resemble the original observations. 
Once trained, these models can produce novel examples that are statistically consistent with the training set, offering a powerful tool for data augmentation, variability exploration of biological variability, and simulation of experimental conditions.

In the context of microbiology, these models can be used to simulate realistic mass spectra, thereby expanding existing MALDI-TOF MS databases, mitigating class imbalance, and enabling controlled in silico experiments. The ability to generate synthetic spectra that mimic the diversity and noise characteristics of real measurements has the potential to strengthen downstream diagnostic algorithms and improve model robustness across acquisition settings.

\subsubsection{Problem Formulation}

Formally, let $\mathbf{x} \in \mathbb{R}^D$ denote a MALDI-TOF spectrum, represented as a vector of $D$ spectral bins corresponding to $m/z$ intensity values. The goal of generative modeling is to learn a model $p_\theta(\mathbf{x})$, parameterized by $\theta$, that approximates the true but unknown data distribution $p_\text{data}(\mathbf{x})$. Once $p_\theta(\mathbf{x})$ is learned, the model can generate synthetic spectra $\tilde{\mathbf{x}} \sim p_\theta(\mathbf{x})$ that resemble real measurements both statistically and visually.

In many practical scenarios, it is desirable to control the generation process so that synthetic samples reflect specific experimental or biological conditions. This motivates the use of conditional generative models, which learn a conditional distribution $p_\theta(\mathbf{x} | c)$ given side information $c$. In this work, $c$ denotes the microbial species associated with each MALDI-TOF spectrum, although it could also represent other clinically relevant metadata such as acquisition batch or strain type.
By sampling $\tilde{\mathbf{x}} \sim p_\theta(\mathbf{x} | c = c^*)$, the model can synthesize spectra corresponding to a desired condition $c^*$ while preserving the intrinsic variability of real data.
In microbiology, this conditional formulation allows for synthesizing spectra from rare species, thereby improving the robustness to class imbalance scenarios.



There are multiple strategies for modeling $p_\theta(\mathbf{x}| c)$, ranging from classical conditional density estimation to modern deep generative models. In this study, we focus on three techniques that have demonstrated strong performance in high-dimensional domains: VAEs \cite{kingma2013auto}, GANs \cite{goodfellow2014generative}, and DDPMs \cite{ho2020denoising,song2020score}.
In the following subsections, we present the conditional formulation of each model and describe the architectural and training adaptations that enable them to capture the distinctive characteristics of MALDI-TOF MS. 

\subsubsection{Conditional Variational Autoencoders}

VAEs \cite{kingma2013auto} are probabilistic latent-variable models that learn a compact representation of high-dimensional data while enabling the generation of new samples. In the context of MALDI-TOF MS, each spectrum $\mathbf{x} \in \mathbb{R}^D$ is mapped into a latent space $\mathbf{z} \in \mathbb{R}^L$ that captures the underlying structure and variability of the spectra. Conceptually, this latent space provides a low-dimensional manifold where spectra with similar characteristics (for example, from the same microbial species) are close to each other, facilitating both controlled generation and interpolation across species.  

A VAE consists of two neural networks: an encoder and a decoder. The encoder, parameterized by $\phi$, approximates the intractable posterior distribution of the latent variables given a spectrum,  
\begin{equation}
    q_\phi(\mathbf{z} | \mathbf{x}) \approx p(\mathbf{z} | \mathbf{x}),
\end{equation}
and outputs the parameters of a multivariate Gaussian in the latent space (mean, $\boldsymbol{\mu}$, and standard deviation, $\boldsymbol{\sigma}$). A latent sample is then obtained via the reparameterization trick \cite{kingma2013auto},  
\begin{equation}
    \mathbf{z} = \boldsymbol{\mu} + \boldsymbol{\sigma} \odot \boldsymbol{\epsilon}, \quad \text{where } \boldsymbol{\epsilon} \sim \mathcal{N}(\mathbf{0}, \mathbf{I}),    
\end{equation}
which allows gradients to propagate through the stochastic layer. The decoder, parameterized by $\theta$, models the likelihood of the data given the latent variable, $p_\theta(\mathbf{x} | \mathbf{z})$, reconstructing the input spectrum from the latent representation.  

In this work, we use the conditional VAE (cVAE) formulation \cite{sohn2015learning}, where both the encoder and decoder are guided by an auxiliary variable $c$ that encodes the microbial species or other metadata associated with each spectrum, having that:
\begin{equation}
    q_\phi(\mathbf{z} | \mathbf{x}, c), \quad p_\theta(\mathbf{x} | \mathbf{z}, c).    
\end{equation}
Conditioning allows the model to learn species-specific latent representations and generate spectra corresponding to a target species $c = c^*$, while maintaining smooth transitions in the latent space.

The model jointly optimizes encoder and decoder parameters by maximizing the conditional evidence lower bound (ELBO):  
\begin{equation}
    \mathcal{L}_{\text{cVAE}}(\theta, \phi; \mathbf{x}, c) = 
\mathbb{E}_{\mathbf{z} \sim q_\phi(\mathbf{z} | \mathbf{x}, c)}[\log p_\theta(\mathbf{x} | \mathbf{z}, c)]
- D_{\mathrm{KL}}(q_\phi(\mathbf{z} | \mathbf{x}, c) \| p(\mathbf{z} | c)).    
\end{equation}

\begin{figure}[t]
    \centering
    \includegraphics[width=\textwidth, page=1, trim={0 2.5cm 0 6.cm},clip]{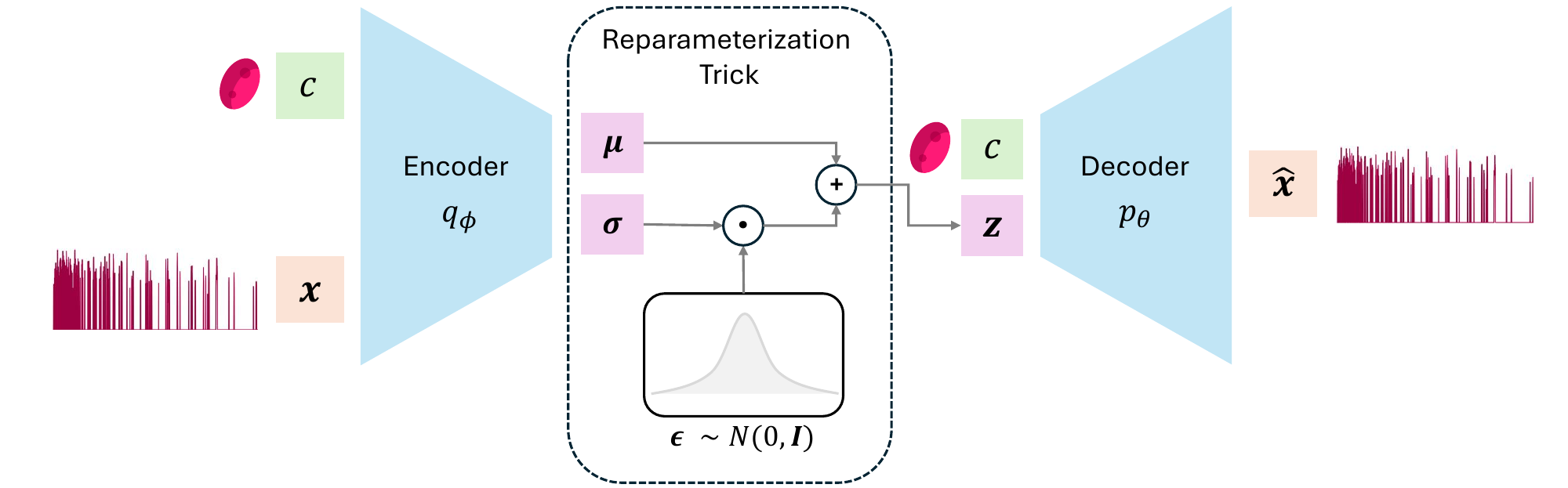}
    \caption{
    Schematic of MALDIVAEs during training. The encoder network compresses the input spectrum $\mathbf{x}$ into a latent vector $\mathbf{z}$. The decoder reconstructs the spectrum $\hat{\mathbf{x}}$ from $\mathbf{z}$, conditioned on microbial species or other metadata $c$. The circle with a dot denotes element-wise multiplication, and the one with a plus sign denotes element-wise addition.
    }
    \label{fig:vae_schema}
\end{figure}

We refer to our adaptation of the formulation of conditional VAEs for MALDI-TOF MS as MALDIVAEs (MALDI-TOF MS Variational Autoencoders), whose training architecture is illustrated in Figure~\ref{fig:vae_schema}.



\begin{figure}[t]
\centering
\includegraphics[width=0.7\textwidth, page=2, trim={5cm 5.5cm 6.cm 5.5cm},clip]{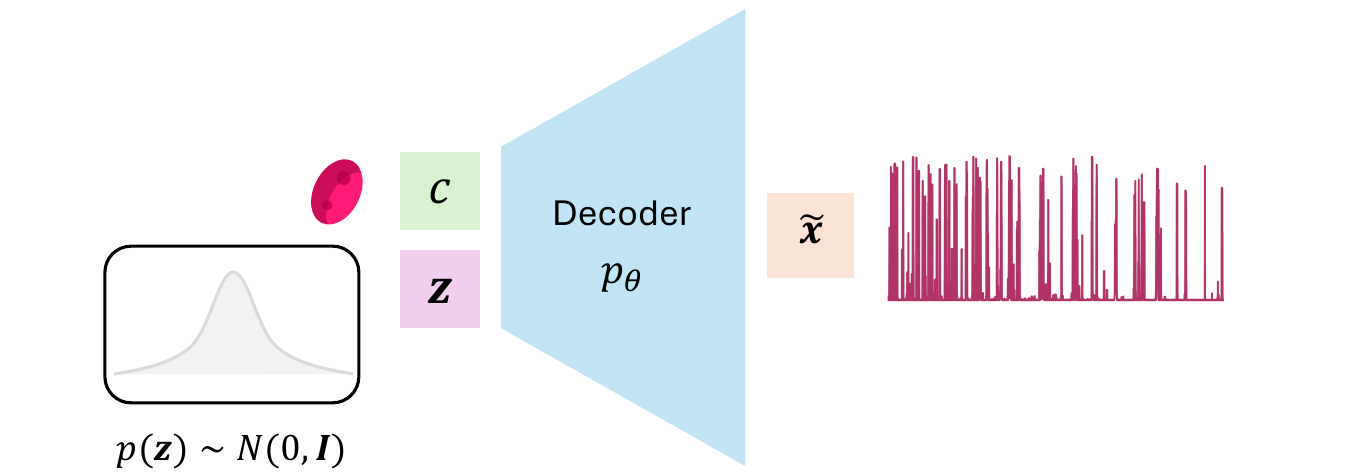}
\caption{
Generative process of MALDIVAEs. New spectra, $\tilde{\mathbf{x}}$ are synthesized by sampling latent variables $\mathbf{z} \sim \mathcal{N}(\mathbf{0}, \mathbf{I})$ and decoding them conditioned on the target label $c$. 
}
\label{fig:vae_generation}
\end{figure}
Once trained, new spectra can be generated by sampling latent variables from the prior and decoding them under a desired condition, that is, $\tilde{\mathbf{x}} = f_\theta(\mathbf{z}, c^*)$ with $\mathbf{z} \sim \mathcal{N}(\mathbf{0}, \mathbf{I})$. Figure \ref{fig:vae_generation} illustrates this generative phase, which complements the training diagram (Figure \ref{fig:vae_schema}). Together, these two figures summarize the dual operation of the MALDIVAE: during training, it learns to encode and reconstruct real spectra while organizing them in a structured latent space; during generation, it exploits this learned structure to synthesize realistic spectra conditioned on any target species $c$. 
 
To effectively model the specific characteristics of MALDI-TOF MS, we adapted the decoder to account for their sparsity and non-negative nature.  
The likelihood $p_\theta(\mathbf{x} |\mathbf{z}, c)$ can, in principle, be modeled using different distributions; however, assuming a Gaussian likelihood led to poor reconstructions, as this distribution fails to capture the asymmetry and sparsity of the signal.  
To address this, since each spectrum was normalized to the range $[0,1]$, the decoder was designed to output parameters of a Soft Bernoulli likelihood, where each bin represents the probability of a non-zero intensity~\cite{vahdat2021score}.  
This formulation effectively models the presence or absence of peaks while preserving differentiability for gradient-based optimization, and in our experiments produced significantly sharper and more realistic peak structures across microbial species compared to other alternatives.


\subsubsection{Conditional Generative Adversarial Networks}

GANs~\cite{goodfellow2014generative} represent a family of generative models that learn to synthesize realistic data through an adversarial training process between two neural networks: a generator and a discriminator.  
The generator $G_\theta$ maps latent variables $\mathbf{z} \sim p(\mathbf{z})$ to synthetic spectra $\tilde{\mathbf{x}} = G_\theta(\mathbf{z}, c)$, while the discriminator $D_\phi$ aims to distinguish between real spectra $\mathbf{x}$ and generated ones $\tilde{\mathbf{x}}$.  
In our formulation, the latent variable $\mathbf{z}$ is drawn from a standard normal prior, $p(\mathbf{z}) = \mathcal{N}(\mathbf{0}, \mathbf{I})$, providing a continuous and smooth latent space from which new spectra can be generated.

In the conditional GAN (cGAN) formulation~\cite{odena2017conditional, mirza2014conditional}, both networks are guided by an auxiliary variable $c$ representing the microbial species or other conditioning information.  
This conditioning enables the model to synthesize spectra corresponding to a particular species and to discriminate between them within that context.  
The resulting objective is: 
\begin{align}
    \min_\theta \max_\phi \;
    &\mathcal{L}_{\text{cGAN}}(\theta, \phi) \nonumber\\
    &=
    \mathbb{E}_{\mathbf{x} \sim p_{\text{data}}(\mathbf{x} | c)} [\log D_\phi(\mathbf{x}, c)] +
    \mathbb{E}_{\mathbf{z} \sim p(\mathbf{z})} [\log (1 - D_\phi(G_\theta(\mathbf{z}, c), c))].    
\end{align}

Through this adversarial interaction, the generator progressively learns to produce synthetic spectra that are indistinguishable from real ones, ultimately approximating the true conditional data distribution $p_{\text{data}}(\mathbf{x} \mid c)$.

\begin{figure}[t]
    \centering
    \includegraphics[width=\textwidth, page=3, trim={0 1.cm 0 3.8cm},clip]{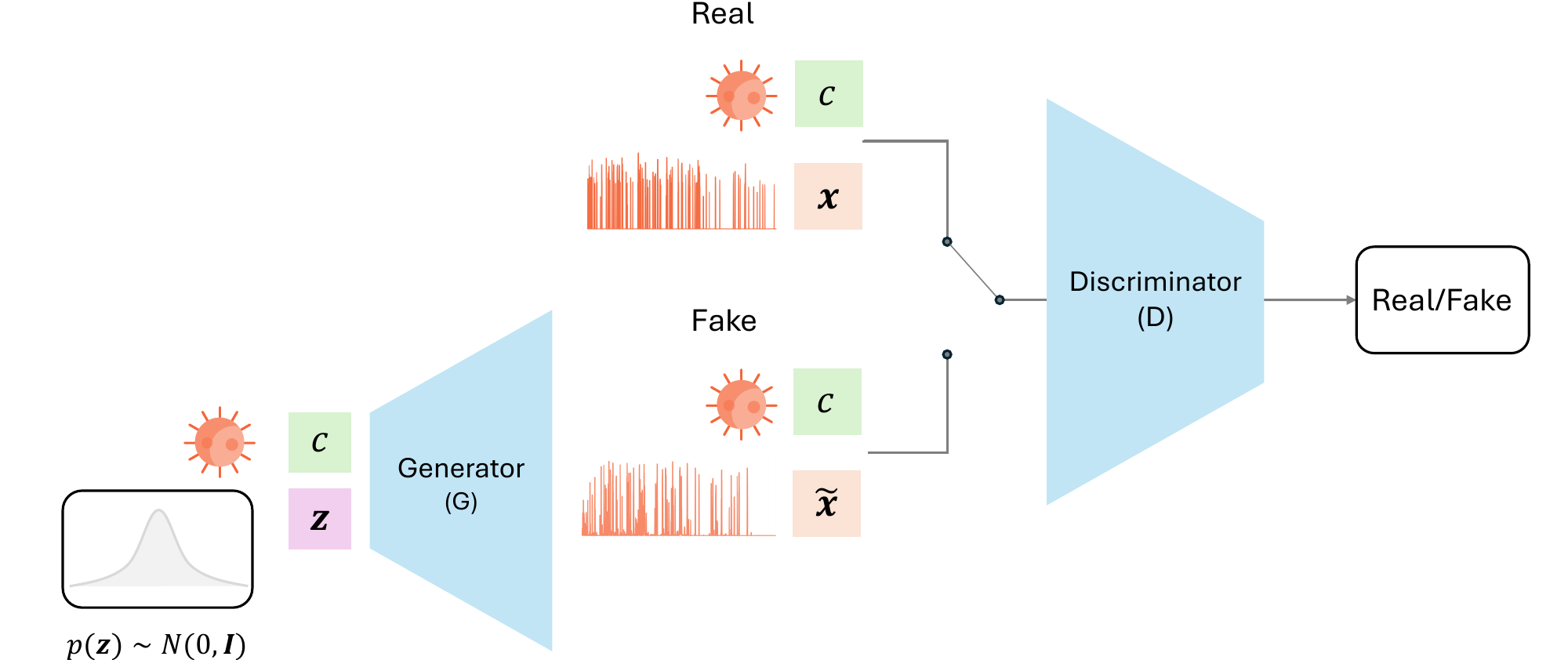}
    \caption{Schematic of MALDIGANs training process. The generator maps a latent vector $\mathbf{z}$ and conditioning label $c$ to a synthetic spectrum $\tilde{\mathbf{x}}$. The discriminator evaluates whether each input spectrum is real or synthetic, taking $c$ into account to enforce class-consistent generation.}
    \label{fig:cgan_schema}
\end{figure}

Figure~\ref{fig:cgan_schema} illustrates the adversarial setup of our conditional GAN adaptation for MALDI-TOF MS, referred to as MALDIGANs (MALDI-TOF MS Generative Adversarial Networks).  
In this configuration, the generator receives a random latent vector $\mathbf{z}$ together with the conditioning label $c$ and learns to produce synthetic spectra that are indistinguishable from real ones, ultimately approximating the true conditional data distribution $p_{\text{data}}(\mathbf{x} \mid c)$.  
The discriminator, in turn, takes real and generated spectra paired with their respective conditions and learns to distinguish genuine from synthetic samples.  
This adversarial training encourages the generator to capture species-specific spectral patterns and reproduce realistic variability rather than global artifacts.


Once trained, MALDIGANs can generate synthetic MALDI-TOF MS for any target species $c$ by sampling latent vectors $\mathbf{z} \sim p(\mathbf{z})$ and decoding them via $\tilde{\mathbf{x}} = G_\theta(\mathbf{z}, c)$. The resulting generative process is equivalent to that of MALDIVAEs (see Figure \ref{fig:vae_generation}), with the decoder replaced by the adversarial generator $G_\theta$ conditioned on both $\mathbf{z}$ and $c$.

To effectively adapt the GAN framework to the characteristics of MALDI-TOF MS, several design modifications were introduced.  
As with MALDIVAEs, to account for the sparsity inherent in MALDI-TOF MS data, the generator outputs a Soft Bernoulli distribution per bin, representing the probability that each bin exhibits non-zero intensity.  
This formulation balances differentiable training with accurate modeling of sparse spectral peaks, allowing the network to reproduce realistic peak distributions.

In addition, during training, we observed that class imbalance across microbial species could lead the generator to overfit majority classes while underperforming on minority ones.  
To mitigate this effect, we incorporated a class-weighted loss that compensates for imbalance by weighting each class inversely to its frequency in the training set.  
This ensures that all species contribute proportionally to the optimization process and that the generator produces representative spectra across the full taxonomic range.

\subsubsection{Conditional Denoising Diffusion Probabilistic Models}

DDPMs \cite{ho2020denoising,song2020score} are a recent type of generative methods that learn to synthesize data by reversing a gradual noising process. Unlike VAEs or GANs, DDPMs explicitly define a stochastic process that transforms real data into noise through a series of small perturbations and then learn to invert this process to recover realistic samples. 

\begin{figure}[!t]
\centering
\includegraphics[width=\textwidth, page=5, trim={3cm 8.cm 3cm 5.cm},clip]{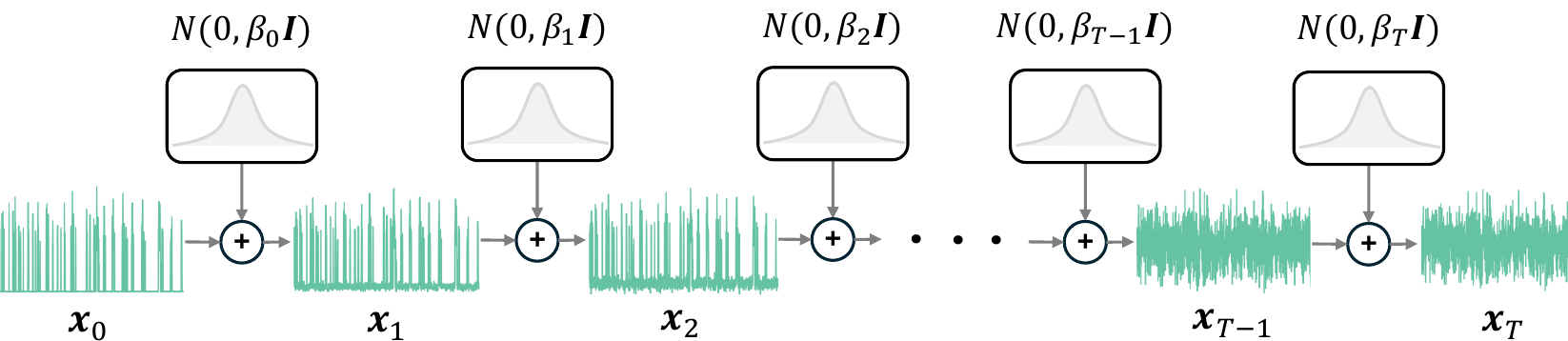}
\caption{Forward diffusion process of MALDIffusion. The real MALDI-TOF spectrum $\mathbf{x}_0$ is gradually degraded into pure noise $\mathbf{x}_T$ through successive noise addition steps. The circle with a plus sign denotes element-wise addition.}
\label{fig:forward_diffusion}
\end{figure}

The forward diffusion process progressively corrupts a real spectrum $\mathbf{x}_0$ over $T$ steps by adding Gaussian noise with increasing variance according to a predefined schedule $\{\beta_t\}_{t=1}^T$. At each step,
\begin{equation}
    q(\mathbf{x}_t | \mathbf{x}_{t-1}) = \mathcal{N}\big(\sqrt{1-\beta_t}\,\mathbf{x}_{t-1}, \beta_t \mathbf{I}\big),    
\end{equation}
resulting in the following cumulative process:
\begin{equation}
    q(\mathbf{x}_t | \mathbf{x}_0) = \mathcal{N}\big(\sqrt{\bar{\alpha}_t}\,\mathbf{x}_0, (1-\bar{\alpha}_t)\mathbf{I}\big),\label{eq:xt_x0}    
\end{equation}
where $\alpha_t = 1 - \beta_t$ and $\bar{\alpha}_t = \prod_{i=1}^t \alpha_i$. As $t$ increases, the signal distribution converges to pure Gaussian noise. Figure~\ref{fig:forward_diffusion} illustrates this process for our adaptation to MALDI-TOF MS data, the MALDIffusion (MALDI-TOF MS Diffusion Model), where spectral peaks gradually vanish as noise accumulates.


\begin{figure}[!t]
\centering
\includegraphics[width=0.8\textwidth, page=7, trim={3.5cm 5.5cm 6cm 5.cm},clip]{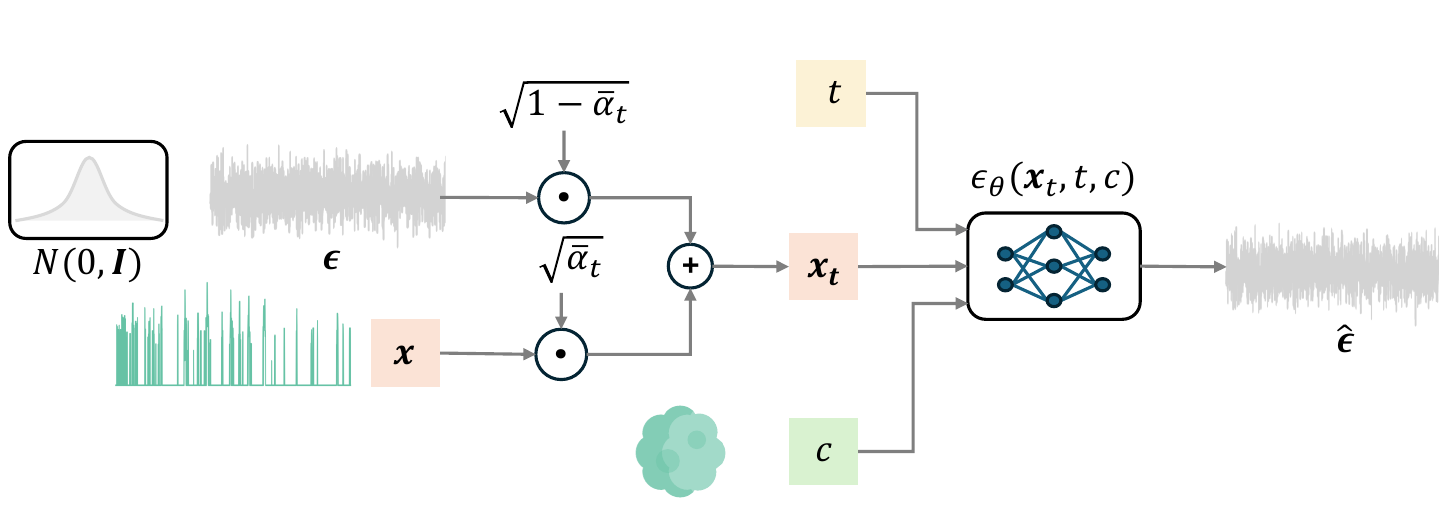}
\caption{Training of the conditional denoiser in MALDIffusion. The network $\boldsymbol{\epsilon}_\theta(\mathbf{x}_t, t, c)$ learns to predict the noise added at each diffusion step, conditioned on the noisy spectrum $\mathbf{x}_t$, the timestep $t$ (sampled uniformly), and the species label $c$. The circle with a dot denotes element-wise multiplication, and the one with a plus sign denotes element-wise addition.}
\label{fig:denoiser_training}
\end{figure}

During training, batches of spectra are sampled from the dataset and corrupted with Gaussian noise according to the forward diffusion process (Figure~\ref{fig:forward_diffusion}).  
For each sample, a diffusion step $t$ is drawn uniformly from $\mathcal{U}\{1, T\}$, and the corresponding noisy spectrum $\mathbf{x}_t$ is computed using the cumulative noising process in Eq.~\eqref{eq:xt_x0}, which combines the original signal scaled by $\sqrt{\bar{\alpha}_t}$ and Gaussian noise scaled by $\sqrt{1-\bar{\alpha}_t}$.  
The denoising network $\boldsymbol{\epsilon}_\theta(\mathbf{x}_t, t, c)$ then learns to predict the noise component $\boldsymbol{\epsilon}$ added at that step, conditioned on the corrupted spectrum, the timestep, and the microbial species.  
Training is performed by minimizing the mean squared error between the true and predicted noise:
\begin{equation}
    \mathcal{L}_{\text{diffusion}}(\theta) = 
    \mathbb{E}_{t, \mathbf{x}_0, \boldsymbol{\epsilon}} 
    \left[\|\boldsymbol{\epsilon} - \boldsymbol{\epsilon}_\theta(\mathbf{x}_t, t, c)\|_2^2\right],
\end{equation}
where $\boldsymbol{\epsilon} \sim \mathcal{N}(\mathbf{0}, \mathbf{I})$.  
Figure~\ref{fig:denoiser_training} illustrates this training process, showing how the network learns to infer the noise added at each diffusion step and progressively recover the structure of the original spectra.

\begin{figure}[t]
\centering
\includegraphics[width=\textwidth, page=10, trim={0 6.cm 0 7.cm},clip]{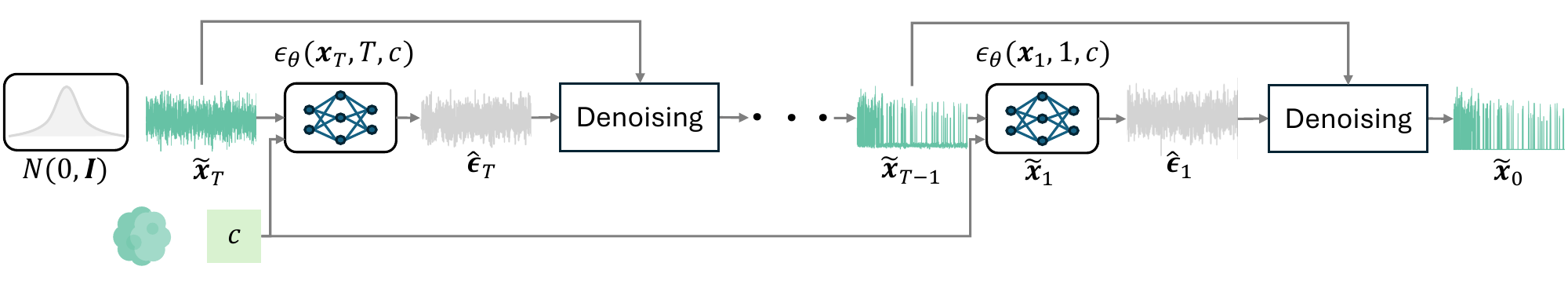}
\caption{Reverse denoising process in MALDIffusion. Generation of synthetic MALDI spectra by iteratively denoising Gaussian noise $\mathbf{x}_T \sim \mathcal{N}(\mathbf{0}, \mathbf{I})$ using the learned model $\boldsymbol{\epsilon}_\theta(\mathbf{x}_t, t, c)$, conditioned on a target microbial species. The denoising block implements Eq. \eqref{eq:denoising} and outputs the denoised MALDI-TOF MS.}
\label{fig:reverse_denoising}
\end{figure}

During generation, the model starts from pure Gaussian noise, $\tilde{\mathbf{x}}_T \sim \mathcal{N}(\mathbf{0}, \mathbf{I})$, and progressively denoises it under the guidance of the microbial species condition $c$.  
At each step $t$, the network refines the noisy spectrum $\tilde{\mathbf{x}}_t$ using the learned denoising function $\boldsymbol{\epsilon}_\theta(\tilde{\mathbf{x}}_t, t, c)$, as illustrated in Figure~\ref{fig:reverse_denoising}, which depicts the iterative reconstruction of a MALDI-TOF spectrum from pure noise.  
The reverse diffusion step is given by:
\begin{equation}
    \tilde{\mathbf{x}}_{t-1} = \frac{1}{\sqrt{\alpha_t}}
    \left(\tilde{\mathbf{x}}_t - 
    \frac{1-\alpha_t}{\sqrt{1-\bar{\alpha}_t}}\,
    \boldsymbol{\epsilon}_\theta(\tilde{\mathbf{x}}_t, t, c)\right)
    + (1-\alpha_t)\mathbf{z},
    \label{eq:denoising}
\end{equation}
where $\mathbf{z} \sim \mathcal{N}(\mathbf{0}, \mathbf{I})$ introduces stochasticity and $\alpha_t$ controls the denoising rate at each step.  
After $T$ iterations, the model produces a synthetic spectrum $\tilde{\mathbf{x}}_0$ consistent with the conditional data distribution $p_\theta(\mathbf{x} \mid c)$, effectively recovering sharp, species-specific peak patterns from pure noise.

To adapt the diffusion framework to MALDI-TOF MS, we implemented a U-Net architecture for the denoising network $\boldsymbol{\epsilon}_\theta(\mathbf{x}_t, t, c)$, specifically modified for one-dimensional spectral inputs.  
Each spectrum was renormalized to the interval $[-1, 1]$ to ensure compatibility with zero-mean Gaussian noise, as is commonly done in image-based diffusion models.  
The conditioning variable $c$ was incorporated into the network through feature-wise modulation layers, allowing the model to control generation in a species-specific manner during both training and sampling.  
These adaptations were essential to ensure stable training and to capture the fine-grained spectral variations characteristic of microbial mass spectra.




\section{Results}
\label{results} 

This section presents a comprehensive evaluation of the proposed generative models. We begin by introducing the quantitative metrics used to assess generation quality. 
Next, the experimental setup is detailed, describing the architectures, training configurations, and hyperparameter choices adopted for each generative model. 

We then analyze the consistency and variability of the spectra generated by MALDIVAEs, MALDIGANs, and MALDIffusions, comparing their ability to capture both the global and fine-grained structures of the training data while maintaining diversity among synthetic samples. A computational analysis follows, examining the complexity, training time, and inference cost associated with each model to evaluate their practical feasibility in real-world microbiology workflows. 

Finally, we assess the utility of the generated data in supervised learning scenarios. Specifically, we conduct experiments to (i) train classifiers exclusively on synthetic spectra and compare their performance against models trained on real data, and (ii) evaluate whether augmenting the training set with generated samples improves classification performance for an underrepresented microbial specie.

All architectures, pretrained models, and code to reproduce the experiments are publicly available at {\small \url{https://github.com/luciaschmidtsantiago/MALDIGen}}.

\subsection{Evaluation Metrics}
\label{sec:metrics}

A key challenge in evaluating generative models for MALDI-TOF MS lies in defining a meaningful notion of similarity between spectra. Unlike continuous signals, MALDI-TOF MS are sparse and irregular, consisting of discrete peaks with varying positions and intensities rather than uniformly sampled values along the $m/z$ axis. Conventional distance measures such as Euclidean or cosine similarity are inadequate in this context, as they are overly sensitive to small peak shifts and fail to capture the localized, non-continuous nature of MALDI-TOF MS.  

To address this, we employ the Peak Information Kernel (PIKE)~\cite{weis2020topological}, a kernel specifically designed for comparing sets of spectral peaks. PIKE represents each spectrum as a collection of weighted impulses and computes a smoothed, shift-tolerant measure of similarity. For two spectra $\mathbf{x}_a$ and $\mathbf{x}_b$ with peak (or bin) positions 
$\{p_{a,i}\}_{i=1}^{D_a}$ and $\{p_{b,j}\}_{j=1}^{D_b}$, 
and corresponding intensities 
$\{\lambda_{a,i}\}_{i=1}^{D_a}$ and $\{\lambda_{b,j}\}_{j=1}^{D_b}$,
the PIKE kernel is defined as:
\begin{equation}
\operatorname{PIKE}(\mathbf{x}_a, \mathbf{x}_b)
= \frac{1}{2\sqrt{2\pi\,t}}
\sum_{i=1}^{n_a}\sum_{j=1}^{n_b}
\lambda_{a,i}\,\lambda_{b,j}\,
\exp\!\left(-\frac{(p_{a,i}-p_{b,j})^2}{8t}\right),
\end{equation}
where $t$ is a bandwidth parameter controlling the tolerance to peak misalignment. In this work, we set $t=8$, which provides sufficient flexibility to account for small variations in peak positions while preserving discriminative power across spectra.
In our implementation, since MALDI-TOF MS were discretized into uniform $m/z$ bins,
we considered $D_a = D_b$ equal to the total number of bins, 
treating each bin as a potential peak location with its associated normalized intensity.
This formulation allows spectra with slightly shifted peaks to remain similar while penalizing larger deviations, making PIKE particularly well-suited to MALDI-TOF MS data that often exhibit acquisition-dependent variability.

We assess the quality of the generated spectra using two complementary groups of metrics:  
(1) \textit{consistency}, which evaluates the fidelity of generated spectra with respect to real samples from the same microbial species, and 
(2) \textit{variability}, which quantifies the diversity of generated spectra and ensures that models do not simply memorize the training data.
To assess consistency, we rely on the Maximum Mean Discrepancy (MMD)~\cite{gretton2012kernel}, while variability is quantified through the Class-Distance (CD) and Neighbour-Distance (ND) metrics, inspired respectively by intra-class diversity~\cite{zhou2020evaluating, heusel2017gans} and nearest-neighbour–based generative evaluation metrics~\cite{sajjadi2018assessing, kynkaanniemi2019improved}. Each metric is adapted using the PIKE kernel to ensure shift-tolerant comparison between spectra.

Given the sets of $N$ real spectra $X = \{x_1,\dots,x_N\}$ and $M$ generated spectra $\tilde{X} = \{\tilde{x}_1,\dots,\tilde{x}_M\}$, the PIKE based squared MMD is defined as:
\begin{align}
\operatorname{MMD}^2(X,\tilde{X}) &=
\frac{1}{N(N-1)} \sum_{n \neq n'} \operatorname{PIKE}(x_n, x_{n'})
+ \frac{1}{M(M-1)} \sum_{m \neq m'} \operatorname{PIKE}(\tilde{x}_m, \tilde{x}_{m'}) \nonumber\\
& \quad - \frac{2}{NM} \sum_{n=1}^{N} \sum_{m=1}^{M} \operatorname{PIKE}(x_n, \tilde{x}_m).
\end{align}
Lower $\operatorname{MMD}^2$ values indicate a closer match between the distributions of generated and real spectra, reflecting higher generative fidelity.

The CD metric quantifies intra-class diversity as the mean pairwise dissimilarity among generated spectra of the same species.  
Let $\tilde{X}_c = \{\tilde{x}_1,\dots,\tilde{x}_{M_c}\}$ be the set of $M_c$ spectra generated for class $c$, then the PIKE based CD metric is defined as:
\begin{align}
\text{CD}(\tilde{X}_c) = \frac{2}{M_c (M_c - 1)} 
\sum_{m < m'} \big( 1 - \operatorname{PIKE}(\tilde{x}_m, \tilde{x}_{m'}) \big),
\end{align}
where larger values indicate greater variability within the generated samples of class $c$.

Finally, the ND metric measures the novelty of generated spectra relative to the training data.  
For each generated spectrum $\tilde{x}_m$, its nearest neighbour $x_m^\ast$ is found in the training set under the PIKE kernel.  
The ND is then defined as the average dissimilarity between each generated spectrum and its closest real counterpart:
\begin{align}
\text{ND}(\tilde{X}) = \frac{1}{M} \sum_{m=1}^M \big( 1 - \operatorname{PIKE}(\tilde{x}_m, x_m^\ast) \big).
\end{align}
Higher ND values indicate that the model produces novel yet biologically plausible spectra, while excessively low values suggest memorization of the training data.

\subsection{Experimental setup}
\label{sec:exp_setup}

All generative models (MALDIVAE, MALDIGAN, and MALDIffusion) were trained exclusively on the training split, and their hyperparameters were selected based on performance on the validation partition.
For the design of the generative models, we explored different network configurations, including MultiLayer Perceptrons (MLP) and one-Dimensional Convolutional Neural Networks (1D-CNN), varying both the number of layers and the dimensionality of the latent space.  
This exploration aimed to identify architectures that balanced generative quality, model stability, and computational efficiency. The conditioning variable $c$, representing the microbial species, was encoded as a one-hot vector for all models, and for MALDIVAE, it was additionally passed through an embedding layer to learn a compact latent representation of the class information.


For both MALDIVAEs and MALDIGANs, we evaluated MLP and 1D-CNN implementations following a common architecture design. All networks consisted of three hidden layers with ReLU activations and were trained using the Adam optimizer with a batch size of 128. Training was conducted for up to 500~epochs for MALDIVAEs and 200~epochs for MALDIGANs, with an early stopping patience of 30~epochs. Different latent space dimensionalities were explored among the set ${8, 32, 64, 128}$. For MALDIVAEs, the learning rate was fixed at $10^{-3}$. In the MALDIGANs, the discriminator and generator used learning rates of $10^{-4}$ and $2\times10^{-4}$, respectively\footnote{A smaller learning rate was used for the discriminator to prevent it from dominating training early, which can lead to generator collapse or convergence to a single class.}. MALDIGANs employed dropout probabilities of 0.2 and 0.3 in the generator and discriminator for the MLP- and 1D-CNN–based implementations, respectively.
In the  1D-CNN configurations, MALDIVAEs used a kernel size of 4 and a max-pooling layer following every two convolutional layers, while MALDIGANs employed a kernel size of 5.

For MALDIffusion, we employed a conditional one-dimensional U-Net $\boldsymbol{\epsilon}_\theta(\mathbf{x}_t, t, c)$ adapted to the spectral domain.  
The architecture progressively reduces the feature dimensionality while increasing the number of channels, evaluating different configurations varying in depth and normalization settings denoted as \textit{S}, \textit{M}, \textit{L}, \textit{XL}, and \textit{Deep}.  
The number of residual blocks per stage was set to two for \textit{S–XL} models and three for the \textit{Deep} configuration.  
Group normalization was applied using 4, 8, 16, and 8 groups for the \textit{S}, \textit{M/L}, \textit{XL}, and \textit{Deep} variants, respectively.  
Additional hyperparameters included a kernel size of 4, a single input channel, and Adam optimization with $\beta_1 = 10^{-4}$ and $\beta_2 = 2\times10^{-2}$.  
Training was performed with a smaller batch size of 64 due to memory constraints, for up to 200~epochs with an early stopping patience of 20~epochs.

From the explored configurations, according to the validation results, we selected three representative models for the subsequent analyses: the 1D-CNN architecture with 8 and 32 latent variables for MALDIVAE and  MALDIGAN, respectively, and the \textit{Deep} variant (3 residual blocks, with 8 groups of 128 channels) of the MALDIffusion.  
These configurations not only provided the best validation performance but also achieved the best trade-off between generative fidelity, diversity, and computational cost. 


\subsection{Comparison of Generative Models}

To evaluate and compare the generative performance of the proposed models, MALDIVAEs, MALDIGANs, and MALDIffusions, we conducted a series of quantitative and qualitative analyses aimed at assessing the fidelity, variability, and perceptual quality of the generated spectra, as well as an evaluation of the complexity and computational cost of the different models. 

For this study, we first assessed the alignment between generated and real spectra using PIKE as a similarity measure.  
For each model and species, the mean PIKE similarity was computed between all generated spectra and 
the complete set of real training spectra belonging to that species (PIKE-all).  
This measure quantifies how closely the generated distributions resemble the real ones at an individual (sample-wise) level.  

In addition, we evaluated the generated spectra using the consistency and variability metrics introduced in Section~\ref{sec:metrics}.  
For each bacterial class, the PIKE-based $\text{MMD}^2$, CD, and ND scores were computed to quantify, respectively, the distributional distance between real and generated spectra, the intra-class diversity, and the novelty with respect to the training data.  
To contextualize the obtained values, each table includes in the first row the corresponding scores computed within the training set itself, providing a reference for the intrinsic intra-class variability of real data for each class.

\begin{table}[!t]
\centering
\caption{
Consistency and variability metrics across models and bacterial species.  
The last column reports the mean $\pm$ standard deviation across species.
}
\label{tab:metrics_pike_all}
\begin{adjustbox}{width=\textwidth}
\begin{tabular}{lccccccccc}
\toprule
\textbf{Metric} & \textbf{Model} & \textit{E. cloacae} & \textit{E. faecium} & \textit{E. coli} & \textit{K. pneumoniae} & \textit{P. aeruginosa} & \textit{S. aureus} & \textbf{Mean $\pm$ SD} \\
\midrule
\multirow{4}{*}{PIKE-all}
 & Baseline & 0.83 $\pm$0.09 & 0.85 $\pm$0.08 & 0.83 $\pm$0.08 & 0.84 $\pm$0.09 & 0.86 $\pm$0.09 & 0.89 $\pm$0.07 & 0.85 $\pm$ 0.02 \\
 & MALDIVAE & 0.89 $\pm$0.06 & 0.91 $\pm$0.06 & 0.88 $\pm$0.06 & 0.90 $\pm$0.05 & 0.91 $\pm$0.07 & 0.92 $\pm$0.05 & 0.90 $\pm$ 0.02 \\
 & MALDIGAN & 0.87 $\pm$0.06 & 0.88 $\pm$0.06 & 0.89 $\pm$0.04 & 0.88 $\pm$0.05 & 0.90 $\pm$0.04 & 0.91 $\pm$0.05 & 0.89 $\pm$ 0.02 \\
 & MALDIffusion & 0.84 $\pm$0.06 & 0.85 $\pm$0.06 & 0.81 $\pm$0.07 & 0.81 $\pm$0.08 & 0.80 $\pm$0.09 & 0.87 $\pm$0.06 & 0.83 $\pm$ 0.02 \\
\midrule
\multirow{4}{*}{MMD$^2$}
 & Baseline & 0 & 0 & 0 & 0 & 0 & 0 & 0 $\pm$ 0 \\
& MALDIVAE  & 0.0062  & 0.0055  & 0.0060  & 0.0045  & 0.0031  & 0.0020 & 0.0045$\pm$0.0016\\
& MALDIGAN  & 0.0484  & 0.0222  & 0.0266  & 0.0170  & 0.0184  & 0.0077 & 0.0234$\pm$0.0123\\
& MALDIffusion  & 0.0571  & 0.0251  & 0.0191  & 0.0311  & 0.0204  & 0.0074 & 0.0267$\pm$0.0150\\
\midrule
\multirow{4}{*}{CD}
 & Baseline & 0.31 $\pm$0.13 & 0.27 $\pm$0.11 & 0.30 $\pm$0.12 & 0.30 $\pm$0.13 & 0.26 $\pm$0.12 & 0.21 $\pm$0.10 & 0.28 $\pm$ 0.04 \\
 & MALDIVAE & 0.26 $\pm$0.12 & 0.22 $\pm$0.09 & 0.27 $\pm$0.12 & 0.26 $\pm$0.12 & 0.22 $\pm$0.12 & 0.18 $\pm$0.09 & 0.23 $\pm$ 0.03 \\
 & MALDIGAN & 0.28 $\pm$0.12 & 0.25 $\pm$0.10 & 0.26 $\pm$0.12 & 0.26 $\pm$0.12 & 0.23 $\pm$0.11 & 0.19 $\pm$0.09 & 0.25 $\pm$ 0.03 \\
 & MALDIffusion & 0.34 $\pm$0.11 & 0.28 $\pm$0.10 & 0.33 $\pm$0.12 & 0.33 $\pm$0.12 & 0.28 $\pm$0.11 & 0.22 $\pm$0.09 & 0.30 $\pm$ 0.05 \\
\midrule
\multirow{4}{*}{ND}
 & Baseline & 0.10 $\pm$0.06 & 0.10 $\pm$0.05 & 0.06 $\pm$0.04 & 0.07 $\pm$0.04 & 0.07 $\pm$0.05 & 0.06 $\pm$0.04 & 0.08 $\pm$ 0.02 \\
 & MALDIVAE & 0.07 $\pm$0.03 & 0.07 $\pm$0.03 & 0.05 $\pm$0.03 & 0.05 $\pm$0.03 & 0.05 $\pm$0.03 & 0.04 $\pm$0.02 & 0.06 $\pm$ 0.01 \\
 & MALDIGAN & 0.07 $\pm$0.03 & 0.10 $\pm$0.03 & 0.05 $\pm$0.02 & 0.05 $\pm$0.02 & 0.05 $\pm$0.02 & 0.05 $\pm$0.02 & 0.06 $\pm$ 0.02 \\
 & MALDIffusion & 0.17 $\pm$0.05 & 0.16 $\pm$0.06 & 0.12 $\pm$0.06 & 0.14 $\pm$0.06 & 0.13 $\pm$0.06 & 0.10 $\pm$0.05 & 0.14 $\pm$ 0.03 \\
\bottomrule
\end{tabular}
\end{adjustbox}
\end{table}

The quantitative results in Table~\ref{tab:metrics_pike_all} summarize the overall performance of the three generative models across all species.  
MALDIVAE achieves the highest PIKE-all values, indicating the closest agreement with real spectra.  
However, when compared to the intrinsic variability of the training set, these high similarities, together with its lower CD and ND values, suggest that MALDIVAE may overfit the dominant spectral patterns, generating highly realistic but less diverse examples.  
MALDIGAN attains slightly lower fidelity but preserves greater intra-class variability, reflecting a better trade-off between realism and diversity.  
MALDIffusion, although showing the lowest PIKE and the highest variability metrics, produces broader and more exploratory generations, partially covering underrepresented regions of the spectral space, but at the cost of reduced accuracy.

The consistency metric ($\text{MMD}^2$) further supports this interpretation: MALDIVAE achieves the smallest distances from the real data distribution, while MALDIGAN and MALDIffusion yield progressively larger values, consistent with their higher diversity and lower reconstruction precision.  
Overall, these results reveal a clear fidelity vs. diversity trade-off among generative techniques, with VAEs excelling in spectral realism and stability, and MALDIGANs and MALDIffusions providing more heterogeneous and exploratory synthesis.


\begin{figure}[t!]
    \centering
    \includegraphics[width=\textwidth]{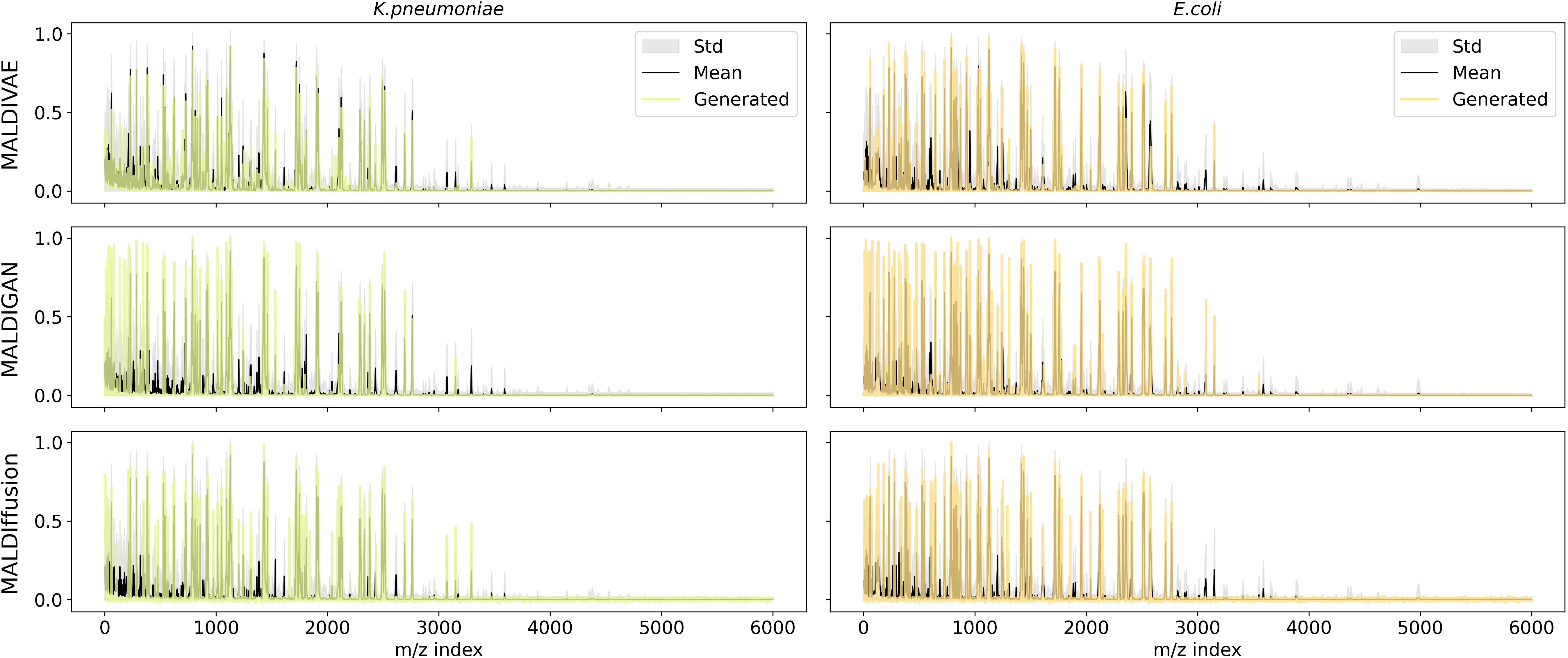}
    \caption{Comparison of real and generated MALDI-TOF MS for two representative species, 
    \textit{K. pneumoniae} and \textit{E. coli}, generated by the three proposed models. 
    For each species and model, the mean real spectrum (\textit{Mean}) and its standard deviation (\textit{Std}) are shown 
    alongside a representative generated spectrum (\textit{Generated}). }
    \label{fig:generated_spectra_models}
\end{figure}

\begin{figure}[thp]
\centering
\begin{subfigure}{0.48\textwidth}
    \centering
    \includegraphics[width=\linewidth]{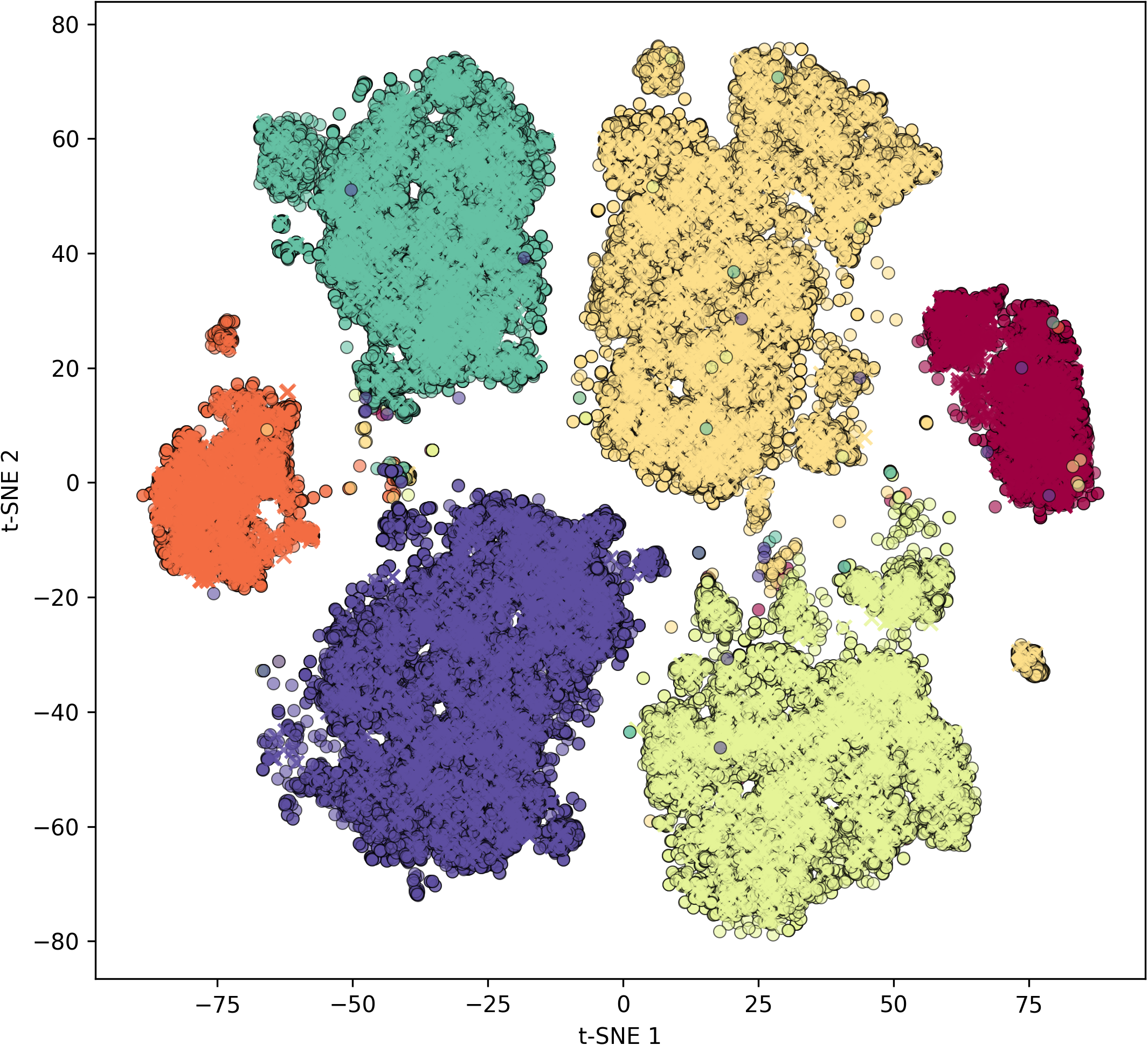}
    \caption{MALDIVAE.}
    \label{fig:tsne_cvae}
\end{subfigure}
\hfill
\begin{subfigure}{0.48\textwidth}
    \centering
    \includegraphics[width=\linewidth]{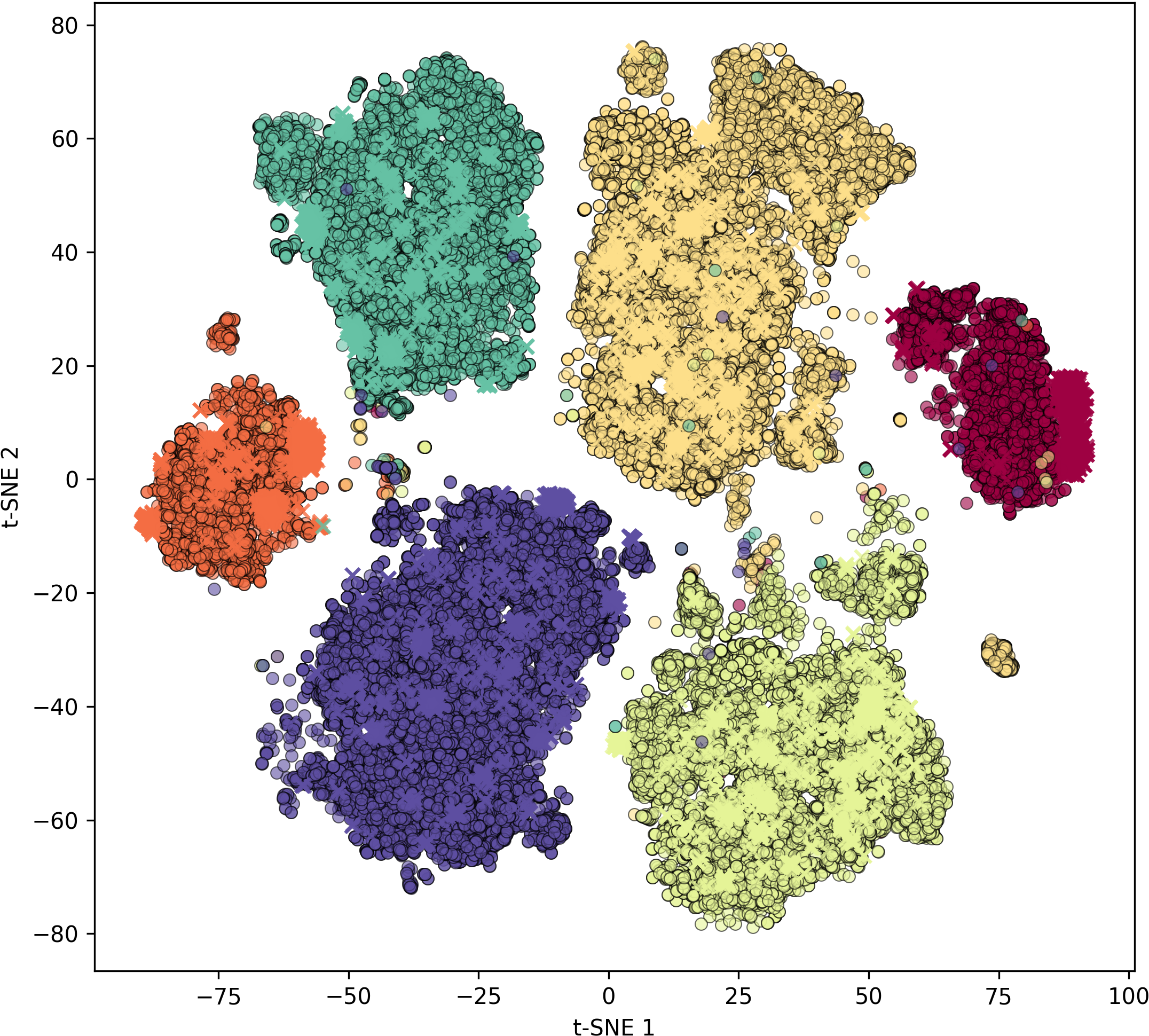}
    \caption{MALDIGAN.}
    \label{fig:tsne_gan}
\end{subfigure}

\vspace{0.25cm}

\begin{subfigure}{0.48\textwidth}
    \centering
    \includegraphics[width=\linewidth]{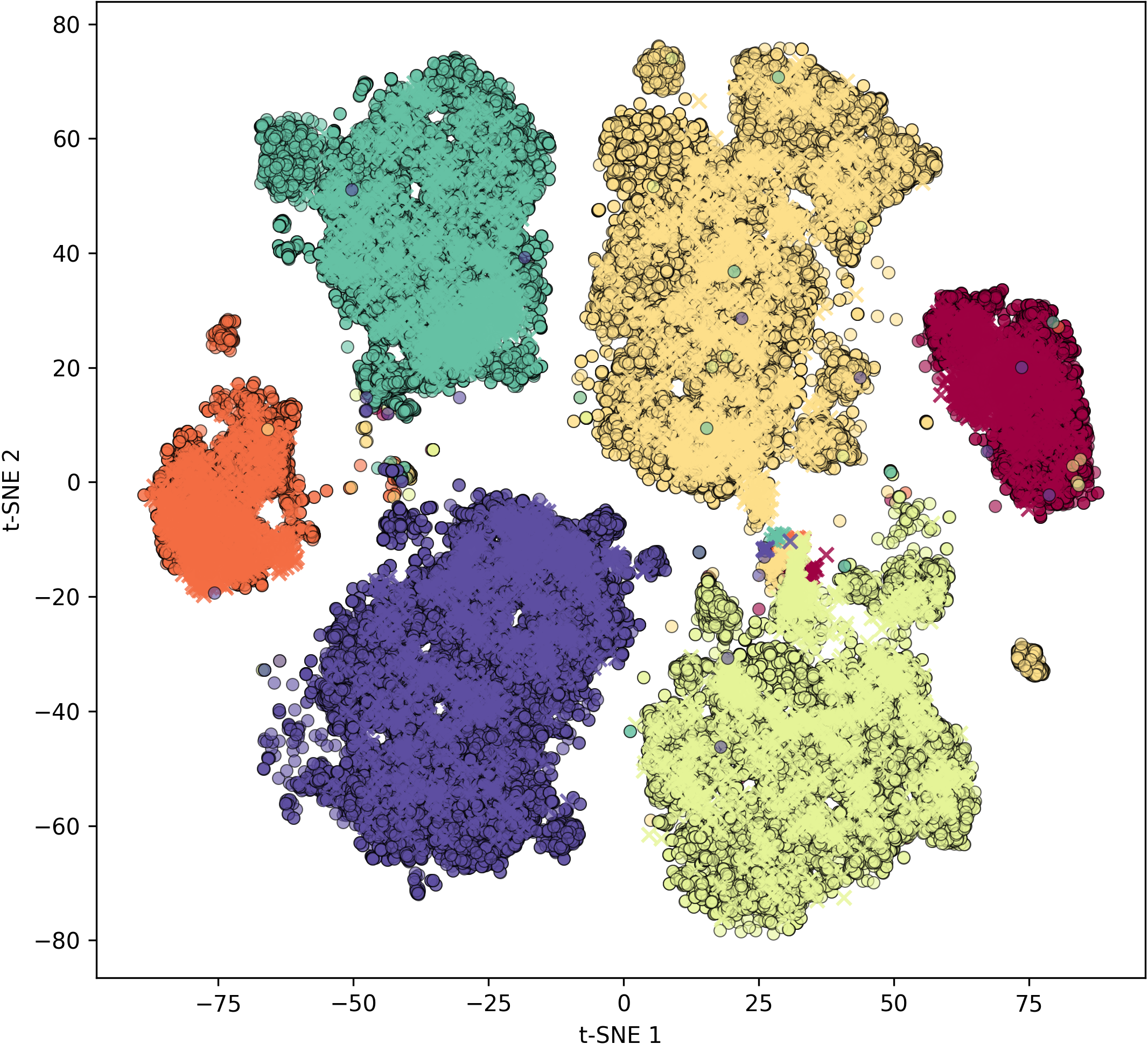}
    \caption{MALDIffusion.}
    \label{fig:tsne_diffusion}
\end{subfigure}
\hfill
\begin{subfigure}{0.45\textwidth}
    \centering
    \includegraphics[width=\linewidth]{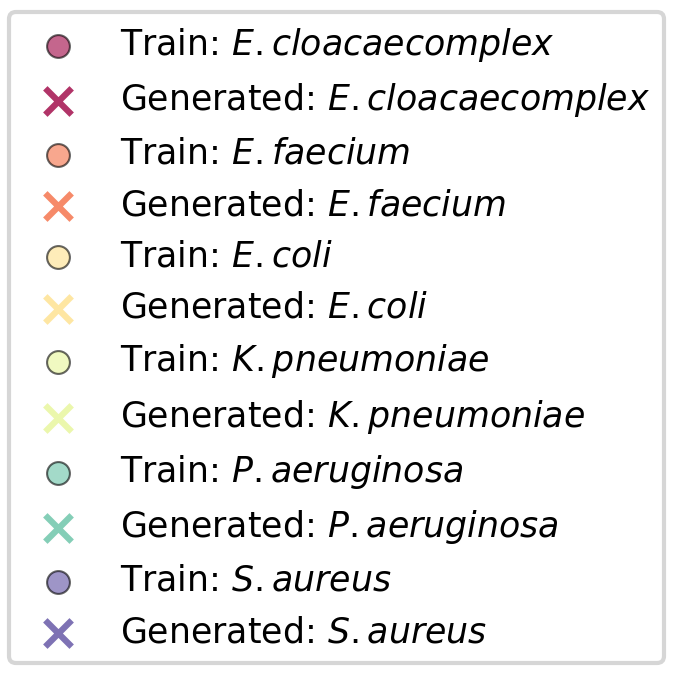}
    \caption{Legend of bacterial species.}
    \label{fig:tsne_legend}
\end{subfigure}

\caption{
t-SNE visualisation of real MALDI-TOF MS alongside synthetic spectra generated by the three conditional generative models. Panel (d) displays the legend indicating species labels and real vs.\ generated data points. The overlap between real and generated clusters illustrates species-level fidelity in the synthetic spectra.
}
\label{fig:tsne_2x2}
\end{figure}

Beyond quantitative metrics, we also performed a qualitative analysis to visually inspect the realism and diversity of the generated spectra.  
For this purpose, Figure~\ref{fig:generated_spectra_models} shows representative examples of generated spectra for each model and two bacterial species (\textit{K. pneumoniae} and \textit{E. coli}). Here, the spectra generated by all three models exhibit realistic peak distributions that closely resemble those observed in real MALDI-TOF MS data.  
The synthetic spectra preserve key spectral characteristics and relative intensity profiles, making them visually indistinguishable from real samples at first glance.  
Among the models, MALDIVAE produces the most coherent peak alignments and intensity patterns, while MALDIGAN and MALDIffusion tend to generate slightly noisier or less regular profiles, though still within biologically plausible ranges.

Besides, Figure~\ref{fig:tsne_2x2} presents two-dimensional t-SNE \cite{maaten2008visualizing} projections combining real training data and synthetic spectra to examine the overlap and dispersion of generated samples within the latent spectral manifold. t-SNE is a nonlinear dimensionality-reduction technique that maps high-dimensional data into a low-dimensional space while preserving local neighbourhood structure. This representation further supports our previous observations.

MALDIVAE achieves the closest reconstruction of the real data distribution, faithfully reproducing even small clusters present in the training set, such as the lower-right \textit{E.~coli} subcluster, indicating strong generalization and class consistency.  
In contrast, MALDIGAN fails to fully cover the distribution, focusing instead on a few dense regions and exhibiting a noticeable shift in the \textit{E.~cloacae} complex, consistent with the large $\text{MMD}^2$ values.  
MALDIffusion performs better in this regard, achieving a broader coverage of the latent space and more uniform distribution than the MALDIGAN, although still not as precise as the MALDIVAE in matching the fine structure of the training data manifold.

Finally, we analyzed the computational complexity of the three models to assess their practical feasibility.  
Table~\ref{tab:computational_cost} summarizes key indicators, including parameter count, training and generation times, and latent dimensionality.  
All models were trained on the same GPU (MSI RTX~5090, 32~GB) under identical conditions for a fair comparison.

\begin{table}[htbp]
\centering
\caption{
Computational cost of the proposed models. The \textit{Bottleneck} column indicates the dimensionality of the latent representation for MALDIVAE and MALDIGAN, and the number of channels at the bottleneck for MALDIffusion. \textit{Train} corresponds to the total training time, \textit{Epoch time} to the average time per epoch, and \textit{Gen} to the average time required to generate a single sample. All times are computed in seconds \textit{(s).}
}
\label{tab:computational_cost}
\begin{adjustbox}{width=\textwidth}
\begin{tabular}{lrrrrrr}
\toprule
Model & Bottleneck & \#Epochs & Train (s) & Epoch time (s) & \#Params & Gen (s) \\
\midrule
MALDIVAE & 8 & 50 & 270.79 & 5.42 & 12.44M & 1.13e-06 \\
MALDIGAN & 32 & 48 & 196.93 & 4.10 & 15.18M & 1.17e-05 \\
MALDIffusion & 128 & 105 & 5911.79 & 56.30 & 24.79M & 0.06 \\
\bottomrule
\end{tabular}
\end{adjustbox}
\end{table}

Both MALDIVAE and MALDIGAN show moderate computational demands, requiring less than five minutes of total training time.  
Although the VAE trains slightly slower due to the reconstruction and regularization terms in its loss, it is the fastest at inference, generating spectra in microseconds.  
MALDIGAN achieves a similar balance of cost and flexibility, with roughly half the epoch time.  

In contrast, MALDIffusion is substantially more demanding: training takes nearly  1 hour and 40 minutes (almost 6,000~s) and sample generation around 0.06~s due to its iterative denoising process, limiting their applicability in real-time or large-scale microbiological settings.  

In summary, across all quantitative and qualitative analyses, MALDIVAE consistently emerges as the most balanced approach, combining strong generative fidelity, realistic spectral diversity, and high computational efficiency.  
MALDIGAN achieves competitive results and captures relevant biological variability, though at the cost of slightly reduced stability and coverage.  
MALDIffusion, while computationally more demanding, yields realistic yet less accurate spectra and does not outperform the other models in reproducing the fine spectral structure observed in real data.  
Together, these results highlight that deep generative models can effectively reproduce the spectral complexity of MALDI-TOF MS data, enabling reliable synthetic data generation for downstream microbiological applications.

\subsection{Classification Performance with Synthetic Data}

To further assess the realism and usability of the generated spectra, we evaluated whether classifiers trained exclusively on synthetic data could achieve performance comparable to those trained on real MALDI-TOF MS.
We used an MLP with two hidden layers (256 and 32 units), ReLU activations, and early stopping (patience of 10 epochs) based on the validation set, with a maximum of 100 epochs.
The classifier was trained and tested on the six major bacterial species analyzed throughout this study, as in the previous section.

Two training configurations were compared:  
(i) an MLP trained using the complete train data partition (real data), and  
(ii) MLPs trained entirely on synthetic spectra generated with the three proposed approaches—MALDIVAE, MALDIGAN, and MALDIffusion—following the same class distribution as in the real dataset.  
Synthetic spectra were produced in equal numbers to the real training samples for each class, ensuring that the original class proportions were preserved.  
Evaluation of all models was carried out on the test and OOD partitions, enabling a robust assessment of generalization across temporal and institutional settings.

\begin{table}[!th]
\centering
\caption{Detection rate (\%) per bacterial species for the test and OOD datasets when using real or synthetic training data.}
\label{tab:detection_results}
\begin{adjustbox}{width=\textwidth}
\begin{tabular}{lcccccccc}
\toprule
\textbf{Dataset} & \textbf{Model} & \textit{E. cloacae} & \textit{E. faecium} & \textit{E. coli} & \textit{K. pneumoniae} & \textit{P. aeruginosa} & \textit{S. aureus} & \textbf{Mean} \\
\midrule
\multirow{4}{*}{\rotatebox{90}{Test}} 
& Real & 98.75 & 99.85 & 99.81 & 99.81 & 99.95 & 99.80 & \textbf{99.68} \\
& MALDIVAE & 94.38 & 98.93 & 99.61 & 99.87 & 100.00 & 99.88 & \textbf{98.78} \\
& MALDIGAN & 95.00 & 99.16 & 99.52 & 99.87 & 99.91 & 99.88 & \textbf{98.89} \\
& MALDIffusion & 81.88 & 99.24 & 99.61 & 99.87 & 99.95 & 99.88 & \textbf{96.74} \\
\midrule
\multirow{4}{*}{\rotatebox{90}{OOD}} 
& Real & 99.84 & 98.11 & 86.16 & 99.60 & 89.43 & 99.73 & \textbf{95.81} \\
& MALDIVAE & 99.53 & 98.86 & 76.90 & 99.80 & 89.15 & 99.83 & \textbf{93.68} \\
& MALDIGAN & 99.53 & 98.86 & 77.14 & 99.80 & 89.43 & 99.97 & \textbf{93.79} \\
& MALDIffusion & 98.89 & 98.86 & 77.41 & 99.80 & 89.01 & 99.93 & \textbf{93.65} \\
\bottomrule
\end{tabular}
\end{adjustbox}
\end{table}

The results in Table~\ref{tab:detection_results} demonstrate that classifiers trained solely on synthetic data achieve performance comparable to those trained on real MALDI-TOF MS.
When analysing performance over the test dataset, detection accuracies across the six bacterial species remain above 96\% for all generative models, with MALDIGAN and MALDIVAE yielding results nearly indistinguishable from the real-data baseline.
When evaluated on the OOD dataset, a moderate decrease in accuracy is observed across all configurations, primarily driven by a substantial drop in performance for \textit{E.~coli}.
This reduction arises from a subset of OOD \textit{E.~coli} spectra corresponding to a distinct strain or subspecies whose spectral distribution differs significantly from the patterns present in the training data.
Consequently, even classifiers trained on real spectra fail to generalize to this subgroup.
Excluding this outlier case, accuracies for the remaining species are highly stable, demonstrating both the robustness and realism of the synthetic spectra and confirming that they can effectively substitute real measurements for model training.

\subsection{Classification of underrepresented species}


In this final experiment, we demonstrate the practical utility of our pretrained generative models in a realistic deployment scenario.  
The aim is to showcase how these models\footnote{All pretrained generative models are also publicly available in the project repository \url{https://github.com/luciaschmidtsantiago/MALDIGen}.} can be exploited by a hospital or laboratory to enrich its own MALDI-TOF dataset, particularly when some species are underrepresented or entirely absent.    
To this end, the generative models are first trained and validated exclusively on the train and validation partitions described earlier, ensuring independence from the data used in this experiment.

Then, to simulate a realistic clinical workflow, we use the OOD-C partition as the hospital’s internal training set, and evaluate generalization on the OOD-D partition, which mimics deployment in a different clinical center with distinct acquisition conditions.  
To make this evaluation representative of real diagnostic settings, we extend the classification task from six to seven species (see Table~\ref{tab:selected_species}), incorporating the minority class \textit{S.~saprophyticus}.  


For the supervised classifier, we employ the same MLP architecture as in the previous section (two hidden layers with 256 and 32 neurons) trained on the OOD-C partition.  
To mitigate the limitations of imbalance and missing species, we additionally train a second MLP using OOD-C augmented with synthetic spectra generated by \textit{MALDIVAE}, \textit{MALDIGAN}, and \textit{MALDIffusion}, balancing each species up to 5{,}000 samples.  
This setup allows us to assess whether synthetic spectra can effectively compensate for local dataset deficiencies and improve generalization across distinct clinical centers.



\begin{figure}[t!]
    \centering
    \begin{subfigure}[t]{0.49\textwidth}
        \centering
        \includegraphics[width=\textwidth, trim={0cm 0.6cm 0cm 0.3cm},clip]{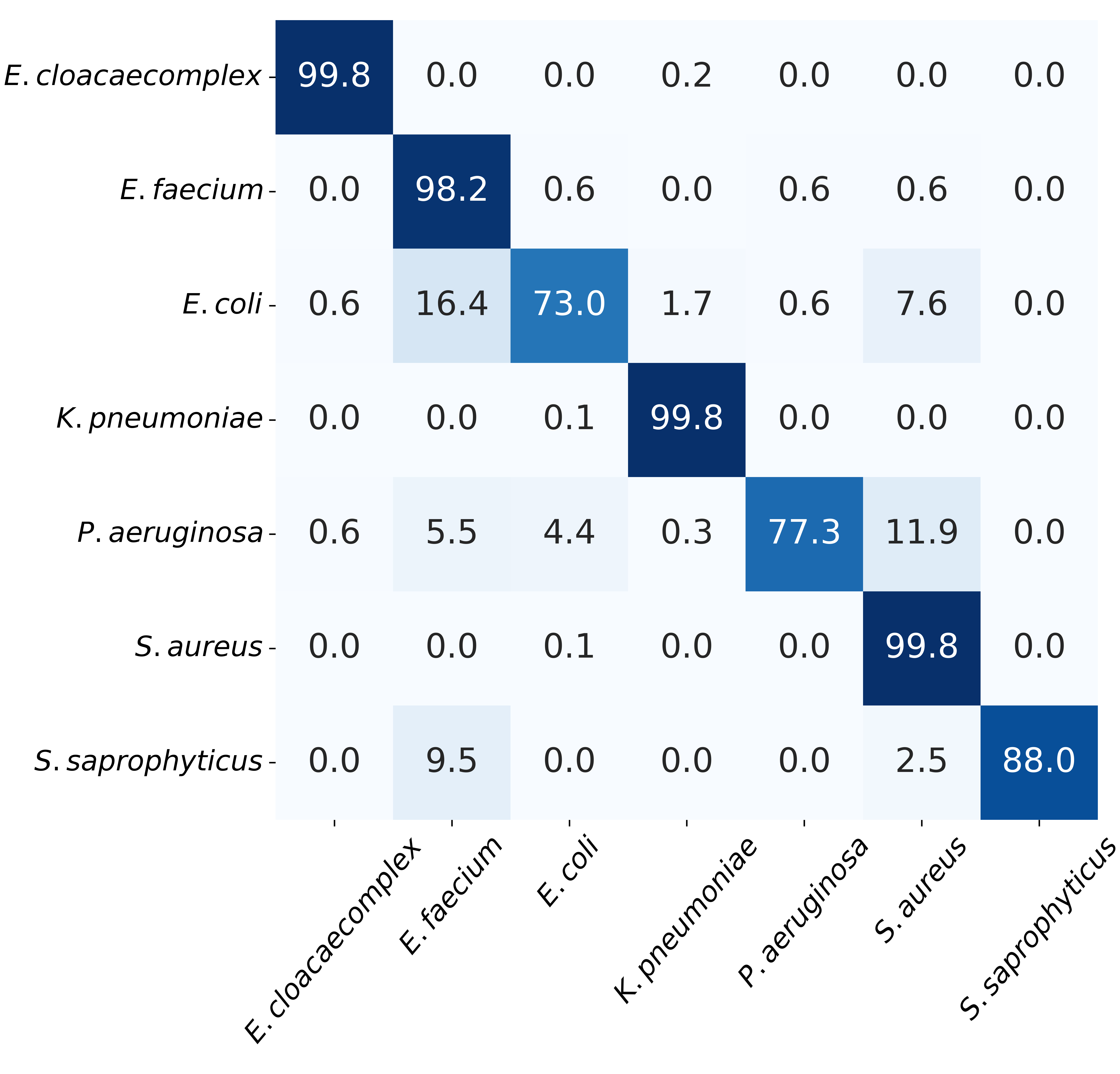}
        \caption{Real only}
        \label{fig:cm_real}
    \end{subfigure}
    \hfill
    \begin{subfigure}[t]{0.49\textwidth}
        \centering
        \includegraphics[width=\textwidth, trim={0cm 0.6cm 0cm 0.3cm},clip]{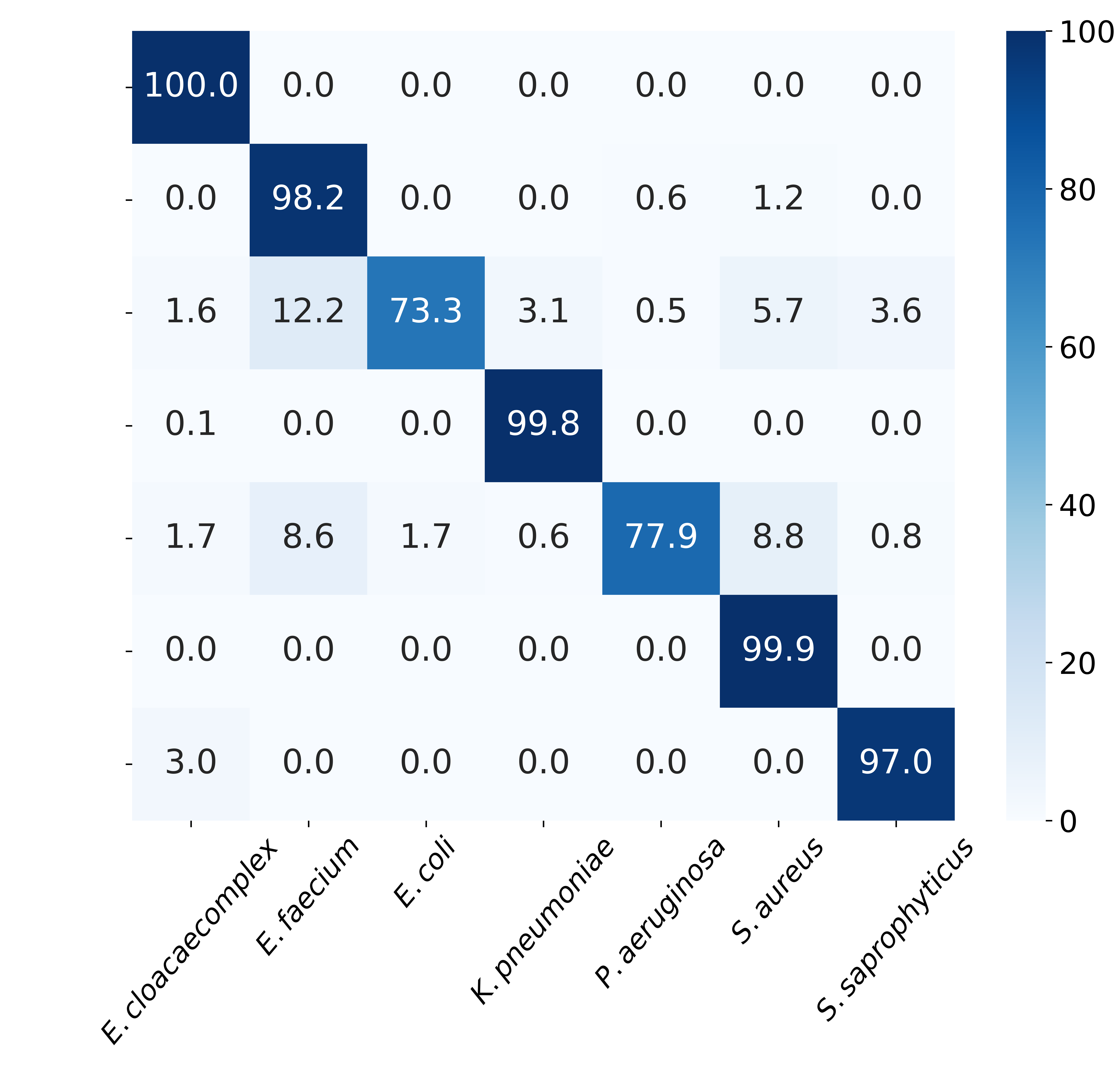}
        \caption{MALDIVAE}
        \label{fig:cm_cvae}
    \end{subfigure}

    \vskip 0.3cm
    \begin{subfigure}[t]{0.49\textwidth}
        \centering
        \includegraphics[width=\textwidth, trim={0cm 0.6cm 0cm 0.3cm},clip]{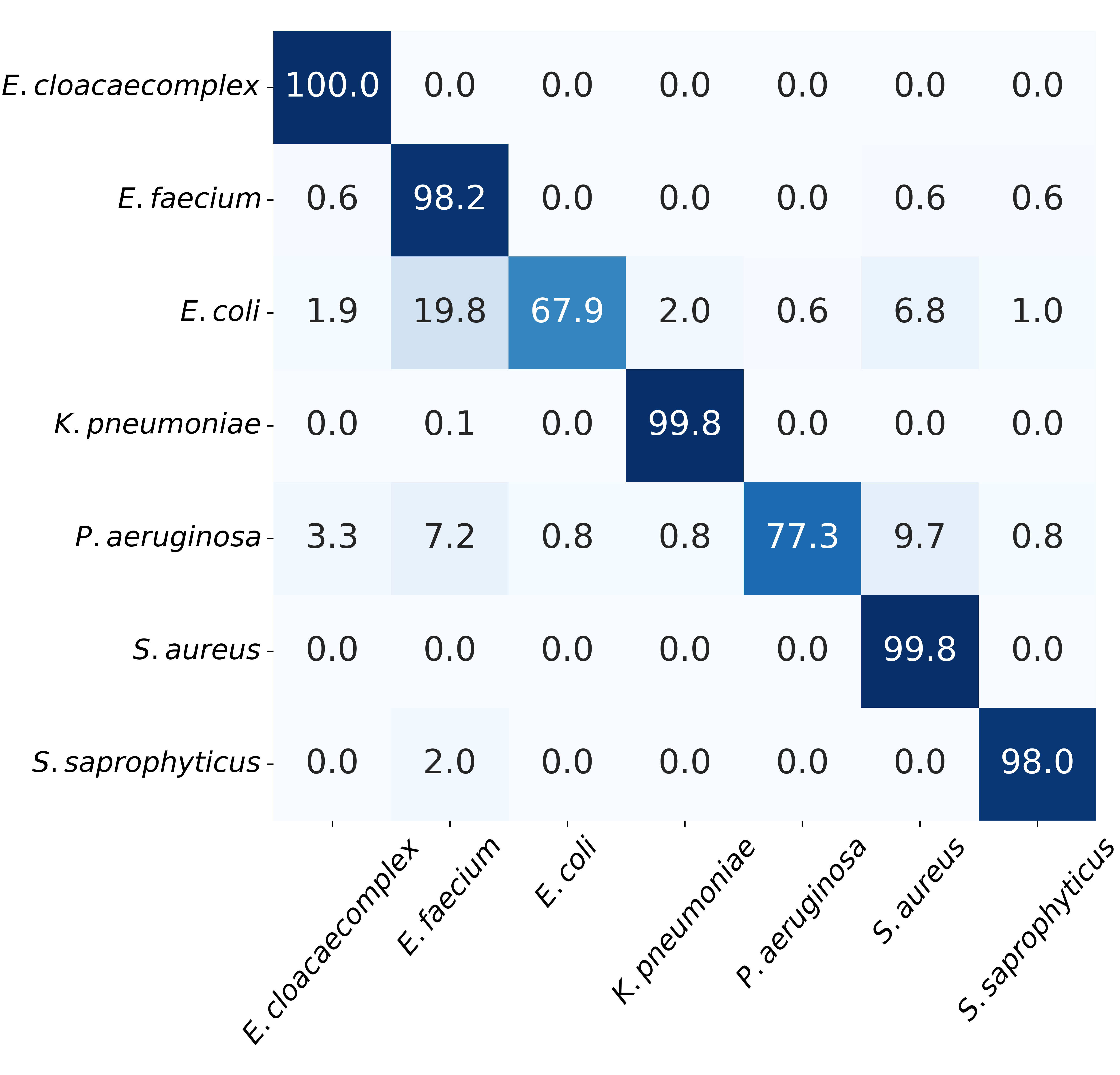}
        \caption{MALDIGAN}
        \label{fig:cm_gan}
    \end{subfigure}
    \hfill
    \begin{subfigure}[t]{0.49\textwidth}
        \centering
        \includegraphics[width=\textwidth, trim={0cm 0.6cm 0cm 0.3cm},clip]{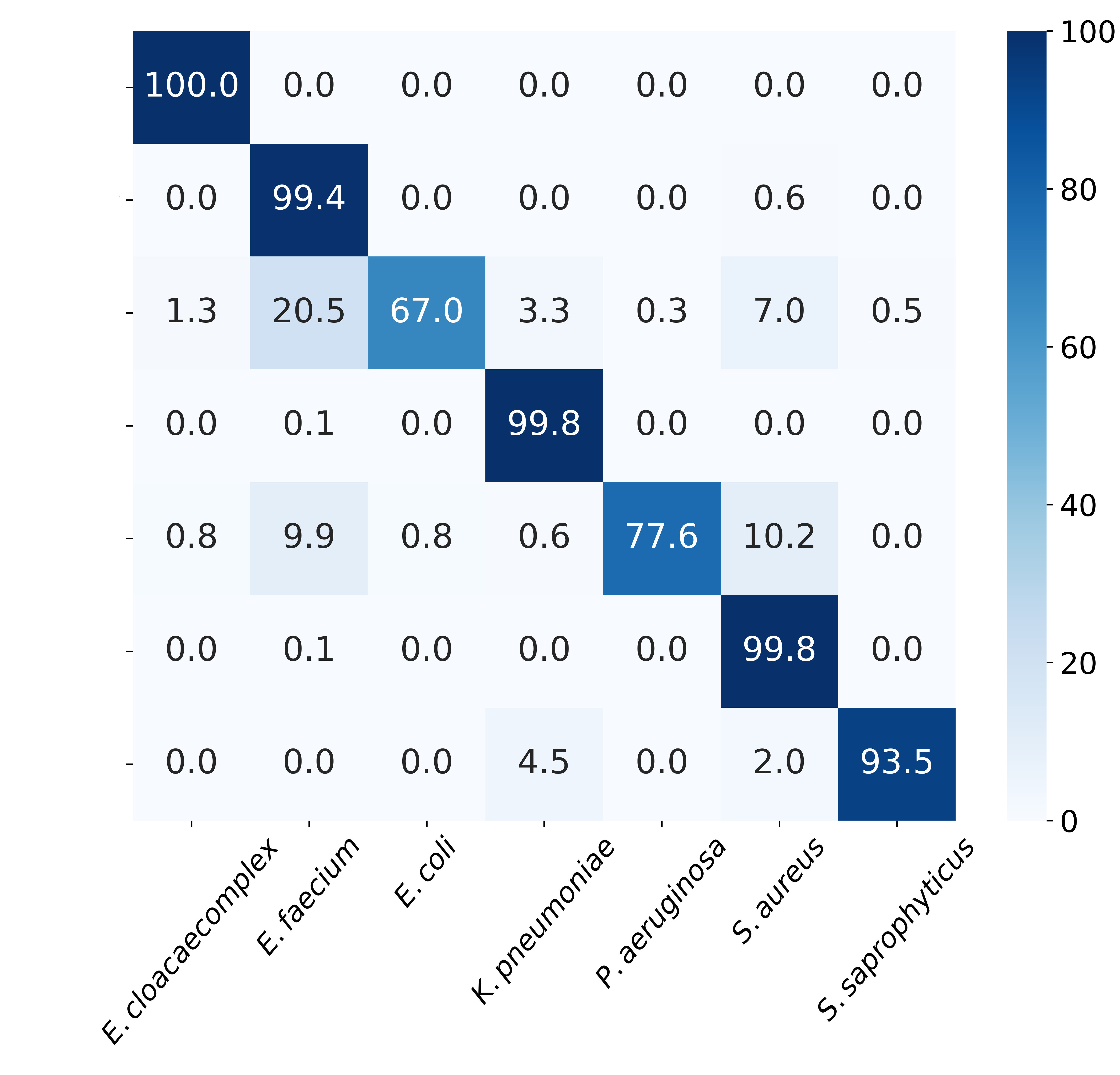}
        \caption{MALDIffusion}
        \label{fig:cm_diffusion}
    \end{subfigure}

    \caption{
        Confusion matrices represented in percentages on the OOD-D dataset for:
        (\subref{fig:cm_real}) the baseline classifier trained only on OOD-C,
        (\subref{fig:cm_cvae}) the model augmented with MALDIVAE,
        (\subref{fig:cm_gan}) the model augmented with MALDIGAN, and
        (\subref{fig:cm_diffusion}) the model augmented with MALDIffusion.
    }
    \label{fig:confusion_matrices}
\end{figure}

Figure~\ref{fig:confusion_matrices} shows the resulting confusion matrices, where each matrix displays the percentage of samples from the true class (rows) predicted as belonging to each class (columns).


Before augmentation, the classifier trained only on OOD-C data performs well for most species, Figure \ref{fig:cm_real}, but exhibits two notable weaknesses: (i) \textit{S.~saprophyticus}, a minority class in the training set, is frequently misclassified, and (ii) \textit{P.~aeruginosa} and \textit{E.~coli} show reduced performance, consistent with observations from the previous section. After balancing the dataset with synthetic spectra generated using {MALDIVAE} (Figure~\ref{fig:cm_cvae}), both limitations are alleviated. The classifier learns a more reliable representation for \textit{S.~saprophyticus}, obtaining a 9\% increase in correctly predicted samples for this class. For \textit{P.~aeruginosa} and \textit{E.~coli}, performance also improves, albeit modestly (0.6\% and 0.3\%, respectively).

Using {MALDIGAN} (Figure~\ref{fig:cm_gan}) further improves the classification of \textit{S.~saprophyticus} by {10\%}, but at the cost of increased errors for \textit{E.~coli}, whose performance drops by {5.1\%}. Finally, {MALDIffusion} (Figure~\ref{fig:cm_diffusion}) exhibits stronger discrimination for the \textit{E.~faecium} class, as indicated by a {1.2\%} performance increase for that species. However, this comes with a {6\% decrease} in \textit{E.~coli} performance and a {5.5\% improvement} for the minority class.

Overall, these results demonstrate that generative augmentation can effectively mitigate class imbalance, although each generative model introduces distinct trade-offs across species.


\section{Conclusions and Future Work}
\label{sec:conclusions}

This work presents an analysis of deep generative models for the synthesis of MALDI-TOF MS in clinical microbiology.  
By adapting three representative paradigms, VAEs, GANs, and DDPMs, to the specific characteristics of microbial spectra, we demonstrated their ability to learn biologically meaningful representations and generate realistic synthetic data.  

Through quantitative metrics, visual inspection, and classification experiments, both the conditional MALDIVAE and MALDIffusion models exhibited high generative fidelity and realistic spectral diversity.  
MALDIGAN delivered competitive performance, though with slightly reduced stability and somewhat narrower distributional coverage.  
However, MALDIffusion required substantially higher computational cost than both MALDIVAE and MALDIGAN, making MALDIVAE the approach that best balances fidelity, diversity, and efficiency for MALDI-TOF MS spectral generation.
Moreover, when incorporated into the training process, synthetic spectra, particularly those generated with {MALDIVAE}, substantially improved the recognition of underrepresented bacterial species, demonstrating that this approach is especially effective for mitigating class imbalance and enhancing model generalization in clinical microbiology.

Through these experiments, we demonstrated that the generated synthetic spectra faithfully reproduce the structure and variability of real MALDI-TOF MS data, to the point of being usable interchangeably with real training samples in downstream analyses. Overall, our findings highlight the potential of generative AI to mitigate class imbalance and enhance the robustness of microbial classification systems.

Future efforts will focus on using the full potential of diffusion-based models, which have become state-of-the-art in other generative domains.  
In particular, we plan to explore more advanced architectures and recently proposed formulations aimed at improving coverage of the spectral distribution, reducing computational cost, and achieving even more precise and detailed reconstructions of microbial spectra.  
In parallel, given the presence of OOD subsets that deviate from the training distribution (e.g., atypical \textit{E.~coli}), we will explore generative approaches for domain alignment and correction of acquisition-induced distribution shifts.
Finally, we will investigate the integration of synthetic data generation with self-supervised learning to build foundational models for MALDI-TOF MS, enabling robust and transferable spectral embeddings across tasks and centers.

\section*{Author contributions}
LSS: conceptualization, data curation, formal analysis, investigation, methodology, software, writing - original draft;
DRT: validation, writing - review \& editing;
CSS: supervision, funding acquisition, investigation, methodology, writing - original draft;
VGV: conceptualization, resources, supervision, funding acquisition, project administration, investigation, methodology, software, writing - original draft.

\section*{Funding sources}

This work is partially supported by grant PID2023-146684NB-I00 funded
by MCIN/AEI/10.13039/501100011033 and ERDF/EU, by the project TEC-2024/COM-89 and intramural project 2023-II-PI-02 funded by Instituto de Investigación Sanitaria Gregorio Marañón (IISGM). The work of CSS was partially funded by UC3M through the Grants for Research Activity by Young Doctors. Funding for APC: Universidad Carlos III de Madrid (Agreement CRUE-
Madroño 2025).

\appendix

\bibliographystyle{elsarticle-num}
\bibliography{bibliography}

\begin{thebibliography}{10}
\expandafter\ifx\csname url\endcsname\relax
  \def\url#1{\texttt{#1}}\fi
\expandafter\ifx\csname urlprefix\endcsname\relax\def\urlprefix{URL }\fi
\expandafter\ifx\csname href\endcsname\relax
  \def\href#1#2{#2} \def\path#1{#1}\fi

\bibitem{dhiman2011performance}
N.~Dhiman, L.~Hall, S.~L. Wohlfiel, S.~P. Buckwalter, N.~L. Wengenack, {Performance and cost analysis of matrix-assisted laser desorption ionization-time of flight mass spectrometry for routine identification of yeast}, J Clin Microbiol 49~(4) (2011) 1614--1616.

\bibitem{patel2013cmr}
R.~Patel, Matrix-assisted laser desorption ionization–time of flight mass spectrometry: A revolution in clinical microbiology, Clinical Microbiology Reviews 26~(3) (2013) 547--603.
\newblock \href {https://doi.org/10.1128/CMR.00072-12} {\path{doi:10.1128/CMR.00072-12}}.

\bibitem{angeletti2017review}
S.~Angeletti, Matrix assisted laser desorption time of flight mass spectrometry (maldi-tof ms) in clinical microbiology: current applications and future perspectives, Clinica Chimica Acta 473 (2017) 1--9.
\newblock \href {https://doi.org/10.1016/j.cca.2016.12.009} {\path{doi:10.1016/j.cca.2016.12.009}}.

\bibitem{popovic2021msr}
N.~Topić~Popovi{\'c}, S.~P. Kazazi{\'c}, I.~Strunjak-Perovi{\'c}, et~al., Sample preparation and culture condition effects on maldi-tof ms identification of bacteria: a review, Mass Spectrometry Reviews 40~(6) (2021) 1589--1603.
\newblock \href {https://doi.org/10.1002/mas.21727} {\path{doi:10.1002/mas.21727}}.

\bibitem{hrabak2013resistance}
J.~Hrab{\'a}k, E.~Chud{\'a}{\v{c}}kov{\'a}, R.~Walkov{\'a}, Matrix-assisted laser desorption ionization–time of flight (maldi-tof) mass spectrometry for detection of antibiotic resistance mechanisms: a review, Journal of Medical Microbiology 62~(11) (2013) 1604--1613.
\newblock \href {https://doi.org/10.1099/jmm.0.058356-0} {\path{doi:10.1099/jmm.0.058356-0}}.

\bibitem{weis2020machine}
C.~V. Weis, C.~R. Jutzeler, K.~Borgwardt, Machine learning for microbial identification and antimicrobial susceptibility testing on maldi-tof mass spectra: a systematic review, Clinical Microbiology and Infection 26~(10) (2020) 1310--1317.
\newblock \href {https://doi.org/10.1016/j.cmi.2020.03.014} {\path{doi:10.1016/j.cmi.2020.03.014}}.

\bibitem{schmidt-santiago2025ml}
L.~Schmidt-Santiago, A.~Guerrero-López, C.~Sevilla-Salcedo, D.~Rodríguez-Temporal, Applied machine learning for human bacteria maldi-tof mass spectrometry: A systematic review, bioRxiv (2025).
\newblock \href {https://doi.org/10.1101/2025.01.25.634879} {\path{doi:10.1101/2025.01.25.634879}}.

\bibitem{Nguyen2024}
H.-A. Nguyen, A.~Y. Peleg, J.~Song, B.~Antony, G.~I. Webb, J.~A. Wisniewski, L.~V. Blakeway, G.~Z. Badoordeen, R.~Theegala, H.~Zisis, et~al., {Predicting Pseudomonas aeruginosa drug resistance using artificial intelligence and clinical MALDI-TOF mass spectra}, {mSystems} 9~(9) (2024) e00789--24.
\newblock \href {https://doi.org/10.1128/msystems.00789-24} {\path{doi:10.1128/msystems.00789-24}}.

\bibitem{deWaele2025pre}
G.~De~Waele, G.~Menschaert, P.~Vandamme, W.~Waegeman, Pre-trained maldi transformers improve maldi-tof ms-based prediction, Computers in Biology and Medicine 186 (2025) 109695.
\newblock \href {https://doi.org/10.1016/j.compbiomed.2025.109695} {\path{doi:10.1016/j.compbiomed.2025.109695}}.

\bibitem{weigand2020driams}
M.~R. Weigand, et~al., Driams: a publicly available database for benchmarking maldi-tof ms-based microbial diagnostics, Scientific Data 7~(1) (2020) 1--8.
\newblock \href {https://doi.org/10.1038/s41597-020-00613-z} {\path{doi:10.1038/s41597-020-00613-z}}.

\bibitem{marisma2025dataset}
D.~Rodríguez-Temporal, L.~Schmidt-Santiago, A.~Guerrero-López, et~al., Marisma: a large-scale maldi-tof database for microbial identification and resistance profiling, \url{https://zenodo.org/records/17201597}, accessed: 2025-11-01 (2025).

\bibitem{brown2020language}
T.~Brown, B.~Mann, N.~Ryder, M.~Subbiah, J.~D. Kaplan, P.~Dhariwal, A.~Neelakantan, P.~Shyam, G.~Sastry, A.~Askell, et~al., Language models are few-shot learners, Advances in neural information processing systems 33 (2020) 1877--1901.

\bibitem{karras2019style}
T.~Karras, S.~Laine, T.~Aila, A style-based generator architecture for generative adversarial networks, in: Proceedings of the IEEE/CVF conference on computer vision and pattern recognition, 2019, pp. 4401--4410.

\bibitem{kingma2013auto}
D.~P. Kingma, M.~Welling, Auto-encoding variational bayes, arXiv preprint arXiv:1312.6114 (2013).

\bibitem{goodfellow2014gan}
I.~Goodfellow, J.~Pouget-Abadie, M.~Mirza, B.~Xu, D.~Warde-Farley, S.~Ozair, A.~Courville, Y.~Bengio, Generative adversarial nets, in: Advances in Neural Information Processing Systems, Vol.~27, Curran Associates, 2014.

\bibitem{mirza2014cgan}
M.~Mirza, S.~Osindero, Conditional generative adversarial nets, arXiv preprint arXiv:1411.1784 (2014).

\bibitem{odena2017conditional}
A.~Odena, C.~Olah, J.~Shlens, Conditional image synthesis with auxiliary classifier {GANs}, in: Proceedings of the 34th International Conference on Machine Learning (ICML 2017), PMLR, 2017, pp. 2642--2651.

\bibitem{ho2020ddpm}
J.~Ho, A.~Jain, P.~Abbeel, Denoising diffusion probabilistic models, in: Advances in Neural Information Processing Systems, Vol.~33, 2020, pp. 6840--6851.

\bibitem{song2020sde}
Y.~Song, J.~Sohl-Dickstein, D.~P. Kingma, A.~Kumar, S.~Ermon, B.~Poole, Score-based generative modeling through stochastic differential equations, arXiv preprint arXiv:2011.13456 (2020).

\bibitem{shorten2019survey}
C.~Shorten, T.~M. Khoshgoftaar, A survey on image data augmentation for deep learning, Journal of Big Data 6~(1) (2019) 60.
\newblock \href {https://doi.org/10.1186/s40537-019-0197-0} {\path{doi:10.1186/s40537-019-0197-0}}.

\bibitem{mittal2023diffusion}
P.~Mittal, C.~Saharia, T.~Salimans, W.~Chan, J.~Ho, M.~Norouzi, Diffusion-based representation learning, in: Proceedings of the 40th International Conference on Machine Learning (ICML 2023), Vol. 202 of Proceedings of Machine Learning Research, PMLR, 2023, pp. 24977--24990.

\bibitem{weis2021driams}
C.~Weis, A.~Cu{\'e}nod, B.~Rieck, K.~Borgwardt, A.~Egli, Driams: database of resistance information on antimicrobials and maldi-tof mass spectra, Dryad, dataset (2021).

\bibitem{schmidt2025marisma}
L.~Schmidt-Santiago, I.~L{\'o}pez-Mareca, M.~Bl{\'a}zquez-S{\'a}nchez, J.~M. Moreno, C.~Sevilla-Salcedo, V.~G{\'o}mez-Verdejo, B.~Rodr{\'\i}guez-S{\'a}nchez, D.~Rodr{\'\i}guez-Temporal, A.~Guerrero-L{\'o}pez, Marisma: a routine maldi-tof ms database from 2018 to 2024, bioRxiv (2025) 2025--05.

\bibitem{goodfellow2014generative}
I.~Goodfellow, J.~Pouget-Abadie, M.~Mirza, B.~Xu, D.~Warde-Farley, S.~Ozair, A.~Courville, Y.~Bengio, Generative adversarial nets, in: NIPS, 2014, pp. 2672--2680.

\bibitem{ho2020denoising}
J.~Ho, A.~Jain, P.~Abbeel, Denoising diffusion probabilistic models, in: NIPS, 2020, pp. 6840--6851.

\bibitem{song2020score}
Y.~Song, J.~Sohl-Dickstein, D.~P. Kingma, A.~Kumar, S.~Ermon, B.~Poole, Score-based generative modeling through stochastic differential equations, arXiv preprint arXiv:2011.13456 (2020).

\bibitem{sohn2015learning}
K.~Sohn, H.~Lee, X.~Yan, Learning structured output representation using deep conditional generative models, in: Advances in Neural Information Processing Systems, Vol.~28, Curran Associates, 2015.

\bibitem{vahdat2021score}
A.~Vahdat, J.~Kautz, Score-based generative modeling in latent space, in: NeurIPS, 2021.

\bibitem{mirza2014conditional}
M.~Mirza, S.~Osindero, Conditional generative adversarial nets, in: arXiv preprint arXiv:1411.1784, 2014.

\bibitem{weis2020topological}
C.~Weis, M.~Horn, B.~Rieck, A.~Cu{\'e}nod, A.~Egli, K.~Borgwardt, Topological and kernel-based microbial phenotype prediction from maldi-tof mass spectra, Bioinformatics 36~(Supplement\_1) (2020) i30--i38.

\bibitem{gretton2012kernel}
A.~Gretton, K.~M. Borgwardt, M.~J. Rasch, B.~Sch{\"o}lkopf, A.~Smola, A kernel two-sample test, The journal of machine learning research 13~(1) (2012) 723--773.

\bibitem{zhou2020evaluating}
Z.~Zhou, W.~Zhang, X.~Ding, X.~Li, Evaluating generative models in the context of multi-class data, arXiv preprint arXiv:2006.10304 (2020).

\bibitem{heusel2017gans}
M.~Heusel, H.~Ramsauer, T.~Unterthiner, B.~Nessler, S.~Hochreiter, Gans trained by a two time-scale update rule converge to a local nash equilibrium, NeurIPS (2017).

\bibitem{sajjadi2018assessing}
M.~S. Sajjadi, O.~Bachem, M.~Lucic, O.~Bousquet, S.~Gelly, Assessing generative models via precision and recall, in: NeurIPS, 2018.

\bibitem{kynkaanniemi2019improved}
T.~Kynkäänniemi, T.~Karras, S.~Laine, J.~Lehtinen, T.~Aila, Improved precision and recall metric for assessing generative models, in: NeurIPS, 2019.

\bibitem{maaten2008visualizing}
L.~v.~d. Maaten, G.~Hinton, Visualizing data using t-sne, Journal of machine learning research 9~(Nov) (2008) 2579--2605.

\end{thebibliography}

\end{document}